\pdfoutput=1
\documentclass{article}
\usepackage[T1]{fontenc}
\usepackage{iclr2027_conference,times}

\usepackage{amsmath,amsfonts,bm}

\def\eqref#1{equation~\ref{#1}}

\def\1{\bm{1}}

\DeclareMathAlphabet{\mathsfit}{\encodingdefault}{\sfdefault}{m}{sl}
\SetMathAlphabet{\mathsfit}{bold}{\encodingdefault}{\sfdefault}{bx}{n}

\usepackage{xcolor}
\usepackage{booktabs}
\usepackage{tabularx}
\usepackage{multirow}
\usepackage{graphicx}
\usepackage{amsmath,amssymb}
\usepackage{bm}
\usepackage{microtype}
\usepackage{subcaption}
\usepackage{enumitem}
\usepackage{xspace}
\usepackage{hyperref}
\usepackage{url}
\usepackage{cleveref}
\usepackage{float}
\usepackage{wrapfig}
\usepackage{tikz}
\usepackage{fontawesome5}
\usetikzlibrary{arrows.meta,positioning,calc}

\newcommand{\referencepreview}{0}
\newcommand{\papercite}[1]{%
  \ifnum\referencepreview=1 [\,]\else\citep{#1}\fi}

\providecommand{\XBenchIconDir}{assets}
\newcommand{\XBenchIcon}[1]{\raisebox{-0.36\height}{\includegraphics[width=1.72em,height=1.72em,keepaspectratio]{\XBenchIconDir/#1.png}}}
\newcommand{\XBenchBest}[1]{\textbf{#1}}

\newcommand{\attacca}{Attacca\xspace}

\definecolor{customcitecolor}{rgb}{0.7, 0.5, 1}

\Crefname{section}{Section}{Sections}
\Crefname{table}{Table}{Tables}
\crefname{section}{Sec.}{Secs.}
\crefname{table}{Tab.}{Tabs.}
\crefname{figure}{Fig.}{Figs.}
\crefname{appendix}{Sec.}{Secs.}
\Crefname{appendix}{Section}{Sections}
\crefname{subappendix}{Sec.}{Secs.}
\Crefname{subappendix}{Section}{Sections}
\crefname{equation}{Eq.}{Eqs.}
\crefname{algorithm}{Algo.}{Algos.}

\hypersetup{
  colorlinks=true,
  citecolor=customcitecolor,
  linkcolor=blue,
  urlcolor=blue
}

    \title{Attacca: Goal-Directed Control under State Continuity for Long-Horizon Embodied Agents}

\author{Gyusik Seo \quad\quad Jaehong Yoon\thanks{corresponding author.} \\
Nanyang Technological University, Singapore \\
\texttt{gyusik001@e.ntu.edu.sg} \quad \texttt{jaehong.yoon@ntu.edu.sg}\vspace{0.2in}\\
\parbox{\textwidth}{\centering
\href{https://attacca-project.github.io}{\faGlobe\ \texttt{https://attacca-project.github.io}}
}
}

\iclrfinalcopy
\begin{document}
\maketitle
\lhead{Preprint}
\setcounter{footnote}{0}
\renewcommand{\thefootnote}{\arabic{footnote}}

\begin{abstract}
A central capability of embodied agents is to accomplish complex objectives through sequences of interdependent tasks. Yet existing visual goal-conditioned policies underlying these agents are typically evaluated on isolated interactions where the target is already visible, and thus do not capture the conditions that arise during continuous long-horizon task execution. In such settings, each task begins from the state left by the previous one: the agent may end at a different position and orientation, the world may have been modified, and the next interaction target may lie outside the current field of view.  As a result, agents relying on such policies may struggle to proceed to the next task when they cannot ground their target in the current observation.
To address this challenge, we propose \attacca{}, a new approach that trains visual goal-conditioned policies on complete search-to-interact trajectories using goal images decoupled from the execution environment.
\attacca{} uses context-decoupled goal sampling to pair each demonstration with a class-compatible masked goal image from another world, removing direct scene and pose correspondence. It learns dense current-view grounding through a target-mask prediction head, providing auxiliary supervision beyond action imitation. We further introduce behavioral-phase conditioning that teaches the policy to distinguish Search, Approach, and Interact stages and adapt its control as execution progresses.
We evaluate \attacca{} on multiple short- and long-horizon embodied tasks in Minecraft. Our method achieves 39.0--47.5\% clean success, improving over the strongest baseline by 1.7--2.4$\times$. On long-horizon tasks, it attains 54\%, 30\%, and 28\% completion, yielding up to a 7$\times$ improvement.

\end{abstract}

\section{Introduction}\label{sec:intro}

Completing long-horizon tasks is a central challenge of embodied intelligence, as each behavior changes the conditions for subsequent ones~\papercite{pmlr-v205-ichter23a,pmlr-v205-huang23c,wang2024voyager,zala2024envgen}. This capability is essential for embodied agents operating in the real world, where accomplishing complex objectives often requires a sequence of interdependent behaviors.
For example, a household robot may need to navigate to a cabinet, retrieve an object, carry it elsewhere, and then locate another object, with each step beginning from the state left by the preceding one.
    
One line of work trains a single policy end-to-end: vision-language-action models map instructions and observations directly to low-level actions. While these methods offer a simple and unified control framework~\papercite{pmlr-v229-zitkovich23a,pmlr-v270-kim25c,octo_2023}, executing a long sequence of interdependent tasks remains challenging for a single policy.
Another line of work adopts hierarchical architectures that pair a high-level reasoner with a low-level policy: the reasoner decomposes a complex objective into subtasks, and the policy executes them sequentially~\papercite{NEURIPS2023_6b8dfb8c,li2025optimus}. 
However, hierarchical decomposition does not by itself resolve the continuity of execution across subtasks. Even with a successful plan, the policy must still carry each task forward from the state produced by the previous one. We refer to this difficulty of carrying execution from one task to the next as the \emph{skill hand-off gap}.

\begin{figure}[ht]
\centering
\includegraphics[width=0.95\textwidth]{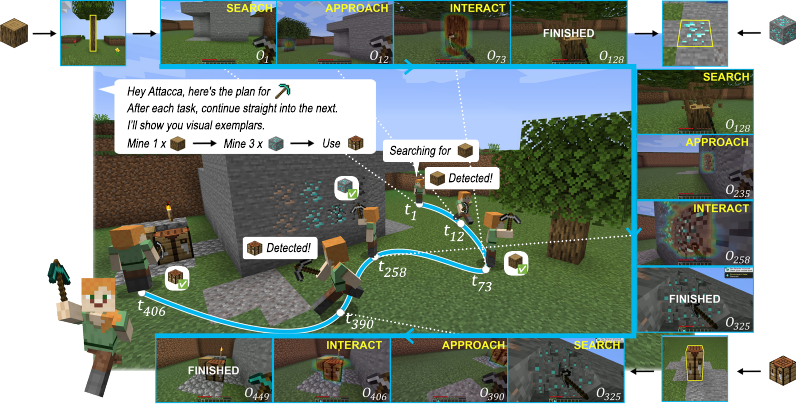}
\caption{\textbf{Goal-directed Control under State Continuity.} The agent executes tasks sequentially, with each task starting from the state left by the previous one. If the target is initially out of view, the agent must search for it, approach it once found, and interact with it to achieve the task.
}
\label{fig:teaser}
\end{figure}

This skill hand-off gap becomes particularly apparent in open-world visuomotor control. We study it in Minecraft, where long-horizon objectives require agents to locate and interact with multiple targets in sequence. 
Recent visuomotor policies, including ROCKET-2~\papercite{cai2025rocket2}, STEVE-1~\citep{NEURIPS2023_dd03f856}, and GROOT~\citep{ICLR2024_16986b69}, have demonstrated strong control capabilities on diverse tasks. However, these methods are largely developed and evaluated on short-horizon tasks, where the agent’s field of view changes only modestly. As a result, robust execution of successive tasks remains underexplored. In particular, completing one task can move the agent into a new state where the next target is not visible, requiring it to locate the target before proceeding.

To address these challenges, we introduce \emph{\attacca{}}\footnote{The name \attacca{} is inspired by \textit{attacca}, the musical direction to continue directly into the next movement without pause.}, a novel approach for learning visuomotor policies that can reliably complete long-horizon tasks through successive task transitions.
\attacca{} trains a target-conditioned policy on search-to-interact trajectories that span the full process of searching for a target when absent, approaching it once discovered, and interacting with it when within reach.
To support continuous and robust target-directed execution, \attacca{} introduces three essential design elements: (1) \textbf{context-decoupled goal sampling}, where each trajectory is paired with a masked target reference from a different environment, allowing the goal to specify what to seek without relying on scene or pose correspondence; (2) \textbf{goal-conditioned target grounding}, where a mask head predicts the target’s visible support over visual patches in the current observation, providing spatial supervision both before and after the target enters the field of view; and (3) \textbf{behavioral-phase conditioning}, where a phase head distinguishes among \textit{Search}, \textit{Approach}, and \textit{Interact} from the current execution state, encouraging the policy to adapt its control as it progresses from target acquisition to interaction.
We jointly optimize action imitation with auxiliary objectives for target grounding and behavioral-phase prediction.

We evaluate \attacca{} across unseen environment configurations, held-out target classes, and target-switching chains of increasing horizon. These chains preserve the state left by each completed task and include transitions where the next target is initially out of view. On the short-horizon embodied tasks Mine, Hunt, and Place, \attacca{} achieves clean success rates of 39.0\%, 47.5\%, and 47.0\%, respectively, outperforming the strongest baseline by 1.7$\times$, 2.4$\times$, and 2.1$\times$. On the long-horizon embodied tasks, it attains 54.0\%, 30.0\%, and 28.0\%, with up to a 7$\times$ gain over the strongest baseline. These results show that \attacca{} remains effective under state continuity across diverse inherited states, previously unseen environments and targets, and longer task horizons, suggesting that jointly learning target acquisition and interaction from environment-decoupled goal images can improve robustness and generalization across targets and task horizons.

\section{Related Work}\label{sec:related}

\paragraph{Target specification in goal-conditioned policies.}
Earlier works defined fixed Minecraft objectives such as obtaining a diamond and trained agents for them~\papercite{ijcai2022p452,hafner2025mastering}, and VPT~\papercite{NEURIPS2022_9c7008af} pre-trained a behavioral prior without a goal input. In goal-conditioned low-level policies, the target can be specified in several modalities. STEVE-1~\papercite{NEURIPS2023_dd03f856} conditions on MineCLIP embeddings~\papercite{fan2022minedojo} of text or of 16-frame behavior clips, and JARVIS-VLA~\papercite{li-etal-2025-jarvis} post-trains a vision-language model for instruction-conditioned control. A text goal can name any target but cannot show what an unfamiliar target looks like, and reported success on atomic tasks remains low~\papercite{NEURIPS2024_85f1225d,pmlr-v267-zheng25j}. GROOT~\papercite{ICLR2024_16986b69} follows reference gameplay videos, which entangle the target with the demonstrator's behavior and the surrounding layout. ROCKET-1~\papercite{Cai_2025_CVPR} and ROCKET-2~\papercite{cai2025rocket2} specify the target with a masked goal image. In their pipelines, a VLM points at the target in the current observation and a segmentation model turns the point into the goal mask~\papercite{Clark_2026_CVPR,ICLR2025_45c1f6a8}. This construction presumes the target is visible, so it is unavailable while the policy is still searching for the target. During training, these controllers also take the goal from the imitated trajectory itself, as in hindsight relabeling~\papercite{NIPS2017_453fadbd,pmlr-v100-lynch20a}, so the goal shares its scene with the observations. We keep the interface of a masked goal image, which clearly designates the target, and decouple it from the execution world. The policy is trained to read what the masked target is from a goal image taken in another world and to find a matching instance in its own surroundings, so the goal remains well-defined even before the target enters the view.

\paragraph{Hierarchical execution and the skill hand-off gap.}
When separately learned skills are composed, one skill can end in a state from which the next cannot succeed, a failure studied in skill chaining~\papercite{NIPS2009_e0cf1f47,lee2019composing,pmlr-v164-lee22a,pmlr-v229-chen23e} and reported as a hand-off problem in mobile manipulation~\papercite{NEURIPS2021_021bbc7e}. Following language-model planners for embodied agents~\papercite{pmlr-v162-huang22a,pmlr-v205-ichter23a,pmlr-v205-huang23c}, hierarchical Minecraft agents such as DEPS~\papercite{NEURIPS2023_6b8dfb8c}, JARVIS-1~\papercite{10778628}, Optimus-1~\papercite{li2025optimus}, and Optimus-2~\papercite{Li_2025_CVPR} dispatch language subgoals from a planner to low-level policies, and DEPS, JARVIS-1, and Optimus-1 identify the controller as a main source of failure on harder tasks. Even with the next subgoal specified, the controller must still acquire the next target from the state left by the preceding skill before it can execute the requested interaction. Some agents acquire targets through simulator state, such as the block queries and pathfinding primitives in Voyager's code skills~\papercite{wang2024voyager} and the lidar observations in Plan4MC's finding skills~\papercite{yuan2023plan4mc}. These systems target long-horizon objectives as a whole, while we study the hand-off within the low-level policy itself, on task chains that preserve the state left by each completed task and require the policy to acquire and interact with the next target from that inherited state.

\section{\attacca{}: Goal-Directed Embodied Control under State Continuity}\label{sec:method}

We introduce \attacca{}, a new approach for training visual goal-conditioned policies to find and interact with targets specified by goal images.
We first formalize goal-directed embodied control under state continuity and describe the policy architecture in~\cref{sec:problem}. We then present the three key components for robust target-directed execution and the joint training objective~(\cref{sec:relabel,sec:grounding,sec:phase}).

\subsection{Problem Statement}\label{sec:problem}
We consider a long-horizon objective specified as an ordered sequence of tasks indexed by \(k\), executed by a low-level policy.
At the start of each task, the agent receives a visual goal $g_k=\left(o_g^k,\,m_g^k\right)$ and an interaction type $c_k$. Here, $o_g^k\in\mathbb{R}^{H\times W\times 3}$ indicates an RGB goal image and $m_g^k\in\{0,1\}^{H\times W\times 1}$ is the goal mask, a binary mask identifying the target in the goal image. 
The goal image serves as a visual reference for what the agent should interact with, showing the target object within a surrounding scene, typically near the center of the image. Because the world encountered at execution time is unknown before exploration, we consider a general setting in which the goal image is sampled in advance from a different world.

We consider two execution regimes: (1) In the single-task setting, the agent receives one visual goal and interaction type, while the corresponding target may initially lie outside its field of view. Success therefore requires the policy to perform the full progression from target search to approach and interaction; (2) In the long-horizon setting, multiple such tasks are executed successively. Each task begins from the state left by the previous one, preserving the agent's pose, camera orientation, and any changes to the world.
As the embodied agent executes a task, it interacts with the environment through low-level actions and receives an egocentric RGB observation $o_t\in\mathbb{R}^{H\times W\times 3}$ at each timestep.
The agent must locate the target, approach it, and execute the requested interaction solely from the visual observation stream and the goal image, without target coordinates, environment state, or external grounding and navigation modules.
For each task, we model the policy at timestep \(t\) as
\begin{equation}\label{eq:policy}
 a_t\sim\pi_\theta\!\left(\,\cdot\mid o_{1:t},\,g_k,\,c_k\right).
\end{equation}
At each timestep, the policy selects a low-level action \(a_t\) based on the observation history \(o_{1:t}\), the current visual goal \(g_k\), and the interaction type \(c_k\). Our goal is to learn a single policy that handles search, approach, and interaction from diverse inherited agent and world states, without relying on scene correspondence between the goal and execution environment.

\begin{figure}[t]
\centering
\includegraphics[width=\textwidth]{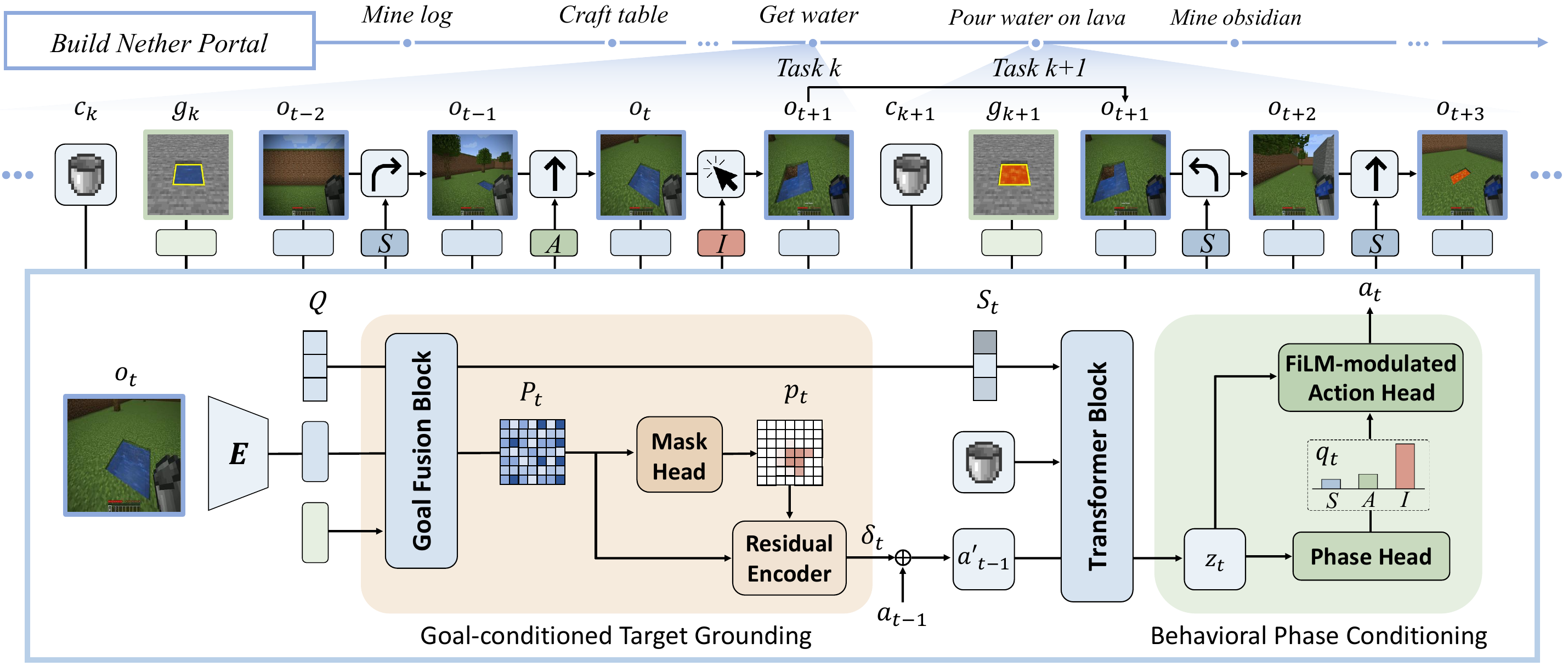}
\caption{Policy architecture of \attacca{}.}
\vspace{-0.1in}
\label{fig:arch}
\end{figure}

\paragraph{Policy architecture.}

\attacca{} is built on a neural policy comprising a visual encoder, a goal-observation fusion module, and a causal Transformer. 
A frozen pretrained visual encoder encodes the goal image once per task and the current observation at each timestep. The goal-fusion module then combines the goal and mask representations with the observation features to emphasize target-relevant visual evidence. The causal Transformer models the fused features over time to predict actions.
With query tokens $Q$ appended to summarize the fused visual context, the module computes
\begin{equation}\label{eq:fusion}
 \left[P_t,\;S_t\right]=\operatorname{GoalFusion}\!\left(\hat o_t,\,\hat g_k,\,Q\right),
\end{equation}
where $\hat o_t$ and $\hat g_k$ denote the patch features of the observation and masked goal, respectively. The outputs at the observation-patch and query positions form the goal-conditioned observation features $P_t = \left[P_{t,1},\cdots,P_{t,N}\right]\in\mathbb{R}^{N\times d}$ and summary tokens $S_t$. 
A causal Transformer processes these tokens with embeddings of the interaction type $c_k$ and the previous action $a_{t-1}$, integrating temporal information across frames.
We use the Transformer output at the previous-action token as the temporal readout $z_t$. After phase modulation~(\cref{sec:phase}), $z'_t$ is mapped by the action head to low-level camera motions and button presses.

\subsection{Context-Decoupled Goal Sampling}\label{sec:relabel}
\begin{figure}[t]
\centering
\includegraphics[width=\textwidth]{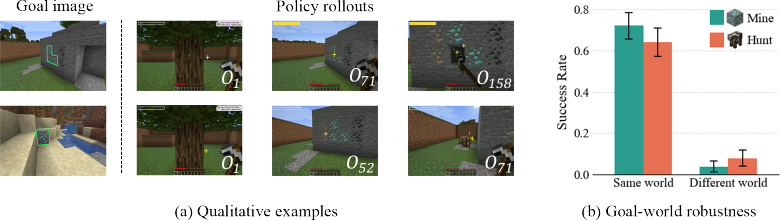}
\caption{Sensitivity to the goal world. (a) Two ROCKET-2 rollouts in the same execution world. (b) Success rates on the mine-diamond-ore and hunt-cow tasks over 50 episodes per condition.}
\vspace{-0.1in}
\label{fig:goal-world}
\end{figure}

In goal-directed embodied control, a reference image specifies the target to be pursued. Existing approaches typically sample this image from the same simulated world in which the demonstration trajectory is collected~\papercite{NEURIPS2023_dd03f856,Cai_2025_CVPR,cai2025rocket2}.
This results in shared visual context between the reference and execution environments, enabling the policy to exploit background and contextual cues as shortcuts even when a target mask is provided. 
Such shortcuts can lead to overfitting and brittle generalization when the target is encountered in unfamiliar environments or visual contexts~(\cref{fig:goal-world}). More importantly, the agent cannot assume that the reference image comes from the same world it will encounter at deployment, since that world is unknown before exploration begins. We therefore decouple goal sampling from the execution context by pairing each trajectory with a reference image of the same target class drawn from a different world.

For each demonstration trajectory $\tau$, let $w(\tau)$ denote the world in which it was collected and $y(\tau)$ its target class, such as an \textit{oak log} or \textit{cow}.
We randomly sample another demonstration of the same target class from a different world, select a frame where the target is visible, and use the corresponding RGB image and target mask as the goal. This ensures
\begin{equation}\label{eq:semantic-contract}
 w(g)\ne w(\tau),\qquad y(g)=y(\tau).
\end{equation}
Since a new goal image is drawn each time a trajectory is sampled, the goal images paired with a demonstration vary in scene but always show the same target class. This design removes direct scene and pose correspondence between the goal image and the demonstration, encouraging the policy to identify matching targets by visual appearance rather than shared scene context. \cref{app:sampling} illustrates the contrast with conventional goal construction and shows consistent gains across goal backgrounds.

\subsection{Goal-Conditioned Target Grounding}\label{sec:grounding}
Without shared scene context, the policy has to locate the target in its own observation. Existing goal-conditioned embodied policies are typically trained via behavior cloning~\citep{NIPS1988_812b4ba2} on demonstration trajectories. While behavior cloning may implicitly encourage target grounding through action prediction, it provides no direct supervision for whether the target is visible or where it appears in the current observation.
To provide such grounding supervision, we introduce a mask head (\cref{fig:arch} middle) that predicts a dense target mask over the goal-conditioned observation patches, indicating the target's visible support in the current observation,
\begin{equation}\label{eq:mask}
 p_{t,i}=\sigma(h_{\mathrm{mask}}(P_{t,i})),\qquad i=1,\ldots,N,
\end{equation}
where $h_{\mathrm{mask}}$ maps each patch feature to a scalar logit, $\sigma$ denotes the sigmoid function, and $p_{t,i}$ is the predicted target-presence probability for patch $i$. 
We derive patch-level supervision from the pixel-level target mask by averaging its values over the region corresponding to each patch,
\begin{equation}\label{eq:mask-target}
 \bar m_{t,i}=\frac{1}{|R_i|}\sum_{(u,v)\in R_i} M_t(u,v),
\end{equation}
where $M_t$ denotes the ground-truth target mask of the current observation and $R_i$ the pixel region corresponding to the $i$-th patch. At each frame, the target loss combines a per-patch binary cross-entropy with a soft Dice loss, where the former provides local patch-wise supervision and the latter encourages spatial overlap between the predicted and ground-truth masks:
\begin{equation}\label{eq:target-loss}
  \mathcal L_{\mathrm{target}}
  =\frac{1}{N}\sum_{i=1}^{N}\operatorname{BCE}\!\left(p_{t,i},\,\bar m_{t,i}\right)
  +\operatorname{Dice}\!\left(p_{t},\,\bar m_{t}\right).
\end{equation}
Frames in which the target is not yet visible receive an all-zero target mask, training the head to capture target visibility as well as localization.
To let the policy act on the predicted mask, a residual encoder aggregates the goal-conditioned features $P_t$ and predicted mask $p_t$ into a single vector $\delta_t$. This vector is added to the readout token before the causal Transformer, so actions can depend  on where the target is predicted to be and whether it is visible. We detach the residual encoder's inputs from the computation graph, so the mask head is trained solely by the target supervision (see \cref{sec:abl-delivery} for a comparison of alternative feedback pathways).

\subsection{Behavioral-Phase Conditioning}\label{sec:phase}
Each task unfolds through three behavioral stages: searching until the target becomes visible, approaching the visible target, and interacting with it at close range. We explicitly model this progression with a behavioral-phase head $h_\phi$ (\cref{fig:arch} right), which predicts the current behavioral phase from the temporal readout $z_t$. The predicted phase then modulates the readout before action prediction.
For supervision, we assign a behavioral-phase label $y_t \in \{\text{Search}, \text{Approach}, \text{Interact}\}$ to each frame based on target visibility and annotated interaction events (\cref{app:dataset}).
The predicted phase distribution conditions the temporal readout through FiLM~\papercite{perez2018film}:
\begin{equation}\label{eq:film}
\begin{aligned}
q_t &= \operatorname{softmax}(h_\phi(z_t)), \\
z'_t &= z_t + W_\beta\operatorname{sg}(q_t) + z_t \odot W_\gamma\operatorname{sg}(q_t),
\end{aligned}
\end{equation}
where $h_\phi$ is a linear head on the readout, $\operatorname{sg}(\cdot)$ denotes stop-gradient, $\odot$ indicates element-wise multiplication, and $W_\beta$ and $W_\gamma$ parameterize learned linear maps. FiLM allows the low-dimensional phase distribution to modulate the readout through feature-wise scaling and shifting, and the resulting $z'_t$ is passed to the action head in place of $z_t$. 
We stop gradients through the phase distribution before FiLM modulation, ensuring that the phase head is trained solely by the phase supervision.
For a sequence of $T$ frames, the behavioral-phase loss and the overall training objective are given by
\begin{equation}\label{eq:full-objective}
\mathcal L_{\phi}
=-\frac{1}{T}\sum_{t=1}^{T}\log q_{t,\,y_t},
\qquad
\mathcal L
=\mathcal L_{\mathrm{BC}}
+\lambda_{\mathrm{target}}\mathcal L_{\mathrm{target}}
+\lambda_\phi\mathcal L_{\phi}.
\end{equation}

We jointly optimize action imitation, target grounding, and behavioral-phase prediction, as given in \cref{eq:full-objective}, where $\lambda_{\mathrm{target}}$ and $\lambda_\phi$ weight the two auxiliary losses. We report their values and a sensitivity sweep in \cref{app:recipe}.
We use ground-truth target masks and phase labels of the current view only for supervision during training, while the policy conditions on its own grounding and phase predictions during both training and inference. At inference time, the policy operates solely on standard inputs without requiring ground-truth annotations or external target information.

\section{Experiments}\label{sec:experiments}

\subsection{Experimental Setup}\label{sec:setup}
\paragraph{Environment.}
We evaluate \attacca{} on goal-directed embodied control in Minecraft. All experiments run in MineStudio~\papercite{cai2024minestudio}. The policy receives egocentric RGB observations and acts in the mouse-and-keyboard action space of VPT~\papercite{NEURIPS2022_9c7008af}. Each task specifies a target class and an interaction type. \attacca{} receives the target as a masked goal image taken in a world other than the execution world, and each baseline receives it in its own goal format~(\cref{fig:goal-spec}).

\paragraph{Training Dataset Construction.} We collect demonstrations recorded by experienced Minecraft players in 1,160 different worlds for the seven training classes of each task. In each demonstration, the player searches for the target, approaches it, and interacts with it, and every frame is labeled with a target mask and a behavioral phase~(\cref{app:dataset}). \attacca{} is trained on these demonstrations, and every baseline other than the released ROCKET-2 checkpoint is trained on the same dataset from its released weights with its native recipe~(\cref{app:baselines}).

\paragraph{Task Design.} \textit{(1) Short-horizon tasks.} Mine, Hunt, and Place each require the agent to find a single target that is initially out of view and to complete the requested interaction, mining it, hunting it, or placing blocks at it, while objects of other classes are present. Each task covers ten target classes, seven seen in training (ID) and three held out (OOD), with 200 episodes per method. We report \emph{clean success}, the fraction of episodes that achieve the goal with no wrong-class interaction, and \emph{interaction precision}, the fraction of an episode's interactions that involve the target class~(\cref{app:benchmark}). \textit{(2) Long-horizon task chains.}
The three chains are Diamond Pickaxe (oak log$\to$\allowbreak diamond ore$\to$\allowbreak crafting table), Wolf Feeding (coal$\to$\allowbreak cow$\to$\allowbreak furnace$\to$\allowbreak wolf), and Nether Portal (water$\to$\allowbreak lava$\to$\allowbreak obsidian$\to$\allowbreak portal$\to$\allowbreak ignition), and each switches to the next goal after every completed stage. The world, position, and camera carry over between stages, so each target must be acquired from the state left by the previous stage. No method is trained on these chains, a shared GUI macro handles cooking and crafting interfaces, and each chain is run with 50 paired policy seeds per method~(\cref{app:helpers}).

\paragraph{Baselines.}
STEVE-1~\papercite{NEURIPS2023_dd03f856} and JARVIS-VLA~\papercite{li-etal-2025-jarvis} receive text goals; STEVE-1 and GROOT~\papercite{ICLR2024_16986b69} accept video as goals; ROCKET-1~\papercite{Cai_2025_CVPR} and ROCKET-2~\papercite{cai2025rocket2} take a current-view goal, a mask built online on the agent's current observation by pointing at the target with Molmo2~\papercite{Clark_2026_CVPR} and segmenting it with SAM~2~\papercite{ICLR2025_45c1f6a8}. 
We also evaluate ROCKET-2 with the masked goal images from other worlds used by \attacca{}, both as the released checkpoint and after fine-tuning on our demonstrations.
See \cref{fig:goal-spec} for examples of each method's goal specification, and \cref{app:baselines} for implementation details.

\subsection{Short-Horizon Task Results}\label{sec:main}
\begin{table}[t]
  \centering
  \caption{Results on short-horizon tasks. ID, OOD, and Avg.\ report clean success, and Prec.\ reports interaction precision. All baselines are trained on our demonstrations. $^{\dagger}$The released ROCKET-2 checkpoint, not trained on our data.}
  \label{tab:short}
  \vspace{-0.1in}
  \small
  \setlength{\tabcolsep}{5pt}
  \renewcommand{\arraystretch}{1.25}
  \resizebox{0.95\textwidth}{!}{\begin{tabular}{@{}l|c| cccc|cccc|cccc@{}}
    \toprule
    \multirow{2}{*}{\textbf{Policy}} & \multirow{2}{*}{\textbf{Goal modality}} & \multicolumn{4}{c|}{\textbf{Mine}} & \multicolumn{4}{c|}{\textbf{Hunt}} & \multicolumn{4}{c}{\textbf{Place}} \\
    & & ID & OOD & Avg. & Prec. & ID & OOD & Avg. & Prec. & ID & OOD & Avg. & Prec. \\
    \midrule
    STEVE-1 & \multirow{2}{*}{Text} & 0.007 & 0.000 & 0.005 & 0.091 & 0.079 & 0.000 & 0.055 & 0.180 & 0.000 & 0.000 & 0.000 & 0.042 \\
    JARVIS-VLA &  & 0.114 & 0.000 & 0.080 & 0.727 & 0.200 & 0.083 & 0.165 & 0.468 & 0.121 & 0.000 & 0.085 & 0.521 \\
    \cmidrule(lr){1-14}
    STEVE-1 & \multirow{2}{*}{Video} & 0.029 & 0.000 & 0.020 & 0.210 & 0.057 & 0.000 & 0.040 & 0.158 & 0.000 & 0.000 & 0.000 & 0.000 \\
    GROOT &  & 0.014 & 0.000 & 0.010 & 0.118 & 0.050 & 0.017 & 0.040 & 0.127 & 0.000 & 0.000 & 0.000 & 0.002 \\
    \cmidrule(lr){1-14}
    ROCKET-1 & \multirow{2}{*}{\begin{tabular}[c]{@{}c@{}}Current view\\$+$ Molmo $+$ SAM\end{tabular}} & 0.021 & 0.017 & 0.020 & 0.237 & 0.186 & 0.167 & 0.180 & 0.324 & 0.057 & 0.017 & 0.045 & 0.081 \\
    ROCKET-2 &  & 0.057 & 0.033 & 0.050 & 0.580 & 0.157 & 0.300 & 0.200 & 0.443 & 0.021 & 0.000 & 0.015 & 0.035 \\
    \cmidrule(lr){1-14}
    ROCKET-2$^{\dagger}$ & \multirow{3}{*}{\begin{tabular}[c]{@{}c@{}}Different world image\\$+$ mask\end{tabular}} & 0.007 & 0.000 & 0.005 & 0.333 & 0.093 & 0.117 & 0.100 & 0.232 & 0.000 & 0.000 & 0.000 & 0.216 \\
    ROCKET-2 &  & 0.264 & 0.167 & 0.235 & 0.739 & 0.193 & 0.167 & 0.185 & 0.558 & 0.286 & 0.083 & 0.225 & 0.567 \\
    \textbf{\attacca{} (Ours)} &  & \XBenchBest{0.450} & \XBenchBest{0.250} & \XBenchBest{0.390} & \XBenchBest{0.778} & \XBenchBest{0.479} & \XBenchBest{0.467} & \XBenchBest{0.475} & \XBenchBest{0.725} & \XBenchBest{0.550} & \XBenchBest{0.283} & \XBenchBest{0.470} & \XBenchBest{0.579} \\
    \bottomrule
  \end{tabular}}
  \vspace{-0.05in}
\end{table}

We first evaluate existing embodied-agent baselines on single tasks across diverse environments, covering both seen and held-out classes. As shown in \cref{tab:short}, STEVE-1 accepts both text and video goals, mapping them into a shared MineCLIP latent goal space. Yet, it is less effective due to limited object-level grounding across environments. JARVIS-VLA receives only a text goal, but benefits from Minecraft-specific visual-language and action post-training, including explicit spatial-grounding supervision, leading to substantially better performance than STEVE-1 and GROOT, particularly on seen classes. ROCKET-1 and ROCKET-2 with a current-view goal obtain a valid goal only when Molmo finds the target in the current observation. While the target is out of view, Molmo returns no point or points at wrong locations~(\cref{app:failure-analysis}), so these pipelines stay at or below 5\% on Mine and Place. On Hunt, animals wander into view on their own, and their success rises to 18--20\%.

Fine-tuning ROCKET-2 on our demonstrations raises its success well above the released checkpoint on every task, which we attribute to learning from the full progression from search to interaction~(\cref{app:dataset}). With the same training data, this masked-goal-image policy also outperforms the text- and video-conditioned baselines on every task, suggesting that a masked goal image is a well-suited interface for goal-directed embodied control. Still, \attacca{}, which uses the same data, initialization, and goals as the fine-tuned ROCKET-2, consistently achieves the highest success rates on both seen and held-out classes. This gap highlights the importance of the proposed context-decoupled goal sampling, target-grounding supervision, and behavioral-phase conditioning.

\subsection{Long-Horizon Task Results}\label{sec:chain-results}
\begin{table}[t]
  \centering
  \caption{Long-horizon task chains: per-stage unconditional success, so the last column of each chain is its chain success rate. $^{\dagger}$The released ROCKET-2 checkpoint, not trained on our data.}
  \label{tab:chains}
  \vspace{-0.1in}
  \small
  \setlength{\tabcolsep}{4.5pt}
  \renewcommand{\arraystretch}{1.25}
  \resizebox{0.95\textwidth}{!}{\begin{tabular}{@{}l|c| ccc|cccc|ccccc@{}}
    \toprule
    \textbf{Policy} & \textbf{Goal modality} & \XBenchIcon{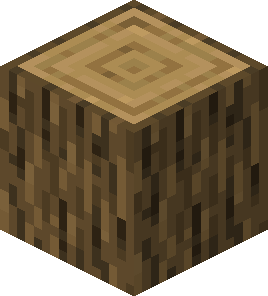} & {\scriptsize$\rightarrow$}\hspace{0.1em}\XBenchIcon{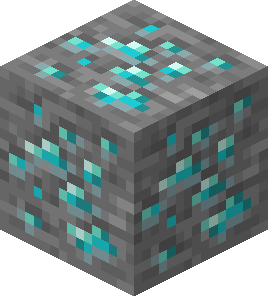} & {\scriptsize$\rightarrow$}\hspace{0.1em}\XBenchIcon{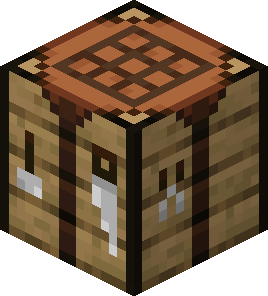} & \XBenchIcon{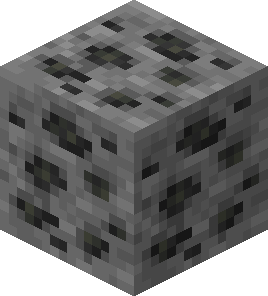} & {\scriptsize$\rightarrow$}\hspace{0.1em}\XBenchIcon{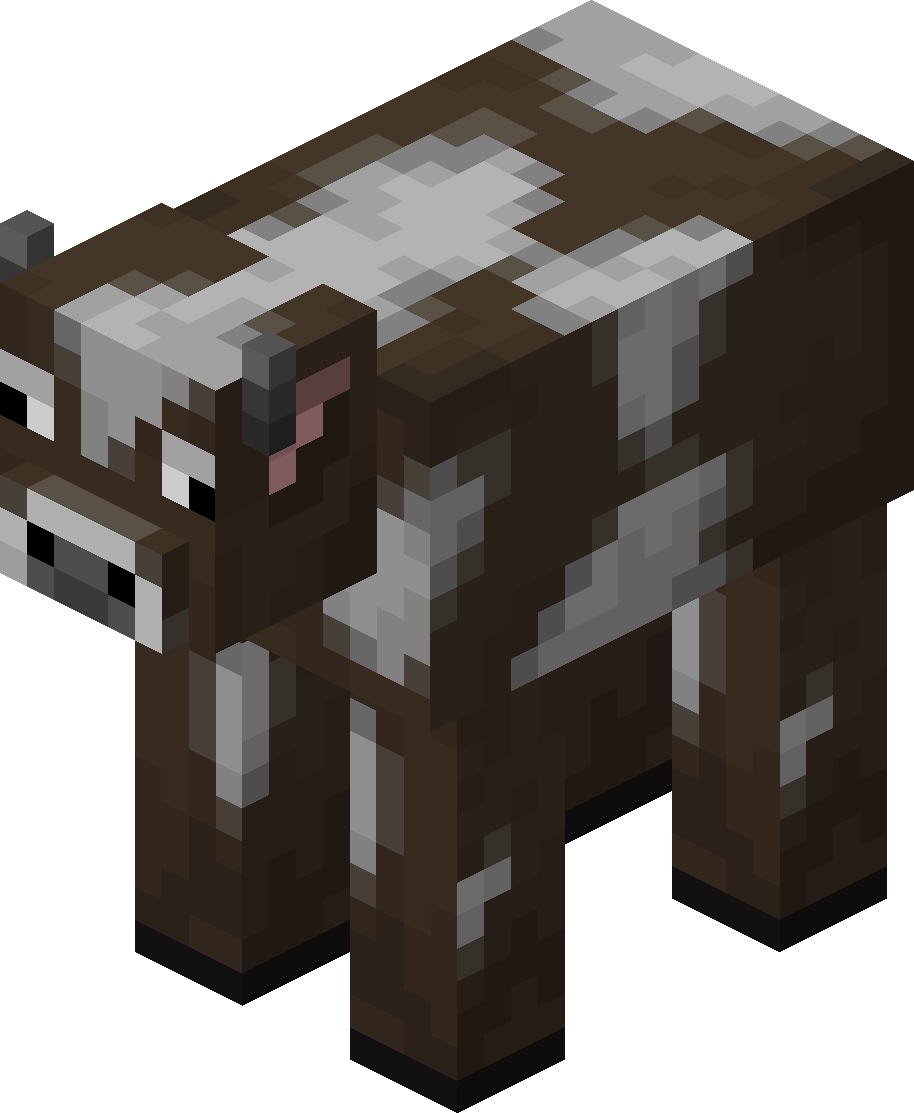} & {\scriptsize$\rightarrow$}\hspace{0.1em}\XBenchIcon{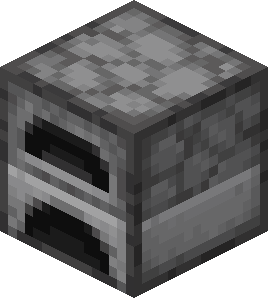} & {\scriptsize$\rightarrow$}\hspace{0.1em}\XBenchIcon{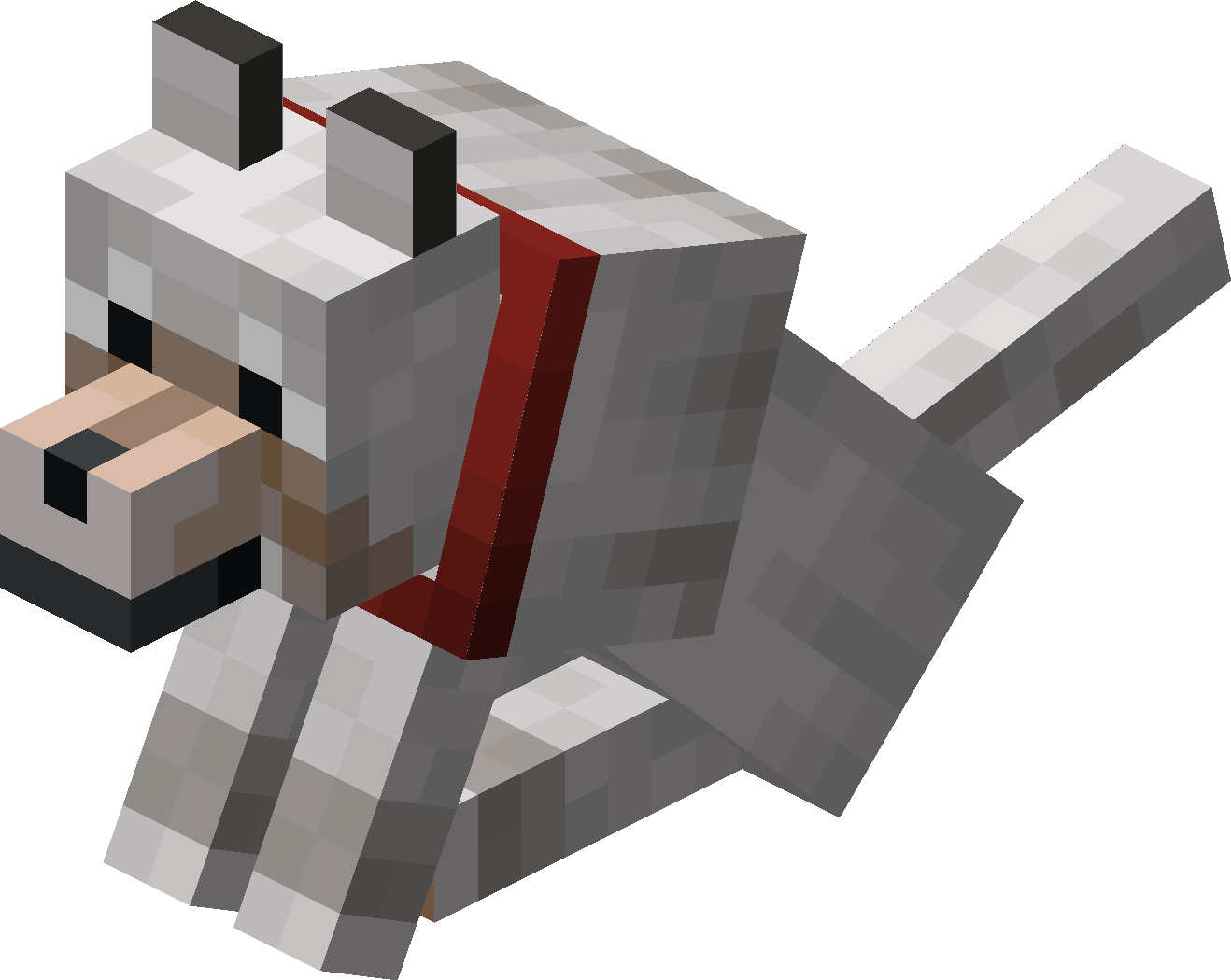} & \XBenchIcon{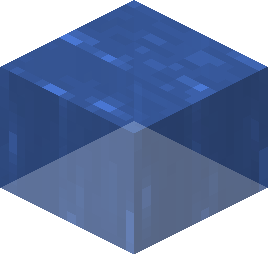} & {\scriptsize$\rightarrow$}\hspace{0.1em}\XBenchIcon{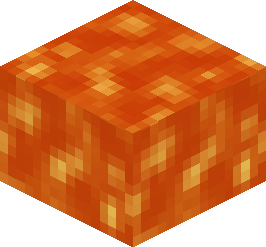} & {\scriptsize$\rightarrow$}\hspace{0.1em}\XBenchIcon{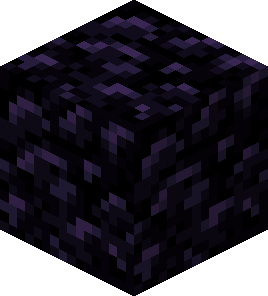} & {\scriptsize$\rightarrow$}\hspace{0.1em}\XBenchIcon{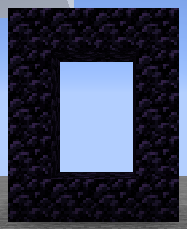} & {\scriptsize$\rightarrow$}\hspace{0.1em}\XBenchIcon{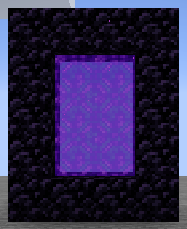} \\
    \midrule
    STEVE-1 & \multirow{2}{*}{Text} & 0.16 & 0.02 & 0.00 & 0.02 & 0.02 & 0.00 & 0.00 & 0.00 & 0.00 & 0.00 & 0.00 & 0.00 \\
    JARVIS-VLA &  & 0.02 & 0.00 & 0.00 & 0.02 & 0.00 & 0.00 & 0.00 & 0.00 & 0.00 & 0.00 & 0.00 & 0.00 \\
    \cmidrule(lr){1-14}
    STEVE-1 & \multirow{2}{*}{Video} & 0.58 & 0.02 & 0.02 & 0.02 & 0.00 & 0.00 & 0.00 & 0.00 & 0.00 & 0.00 & 0.00 & 0.00 \\
    GROOT &  & 0.06 & 0.00 & 0.00 & 0.00 & 0.00 & 0.00 & 0.00 & 0.10 & 0.00 & 0.00 & 0.00 & 0.00 \\
    \cmidrule(lr){1-14}
    ROCKET-1 & \multirow{2}{*}{\begin{tabular}[c]{@{}c@{}}Current view\\$+$ Molmo $+$ SAM\end{tabular}} & 0.00 & 0.00 & 0.00 & 0.02 & 0.00 & 0.00 & 0.00 & 0.06 & 0.00 & 0.00 & 0.00 & 0.00 \\
    ROCKET-2 &  & 0.06 & 0.04 & 0.00 & 0.20 & 0.04 & 0.00 & 0.00 & 0.18 & 0.02 & 0.02 & 0.00 & 0.00 \\
    \cmidrule(lr){1-14}
    ROCKET-2$^{\dagger}$ & \multirow{3}{*}{\begin{tabular}[c]{@{}c@{}}Different world image\\$+$ mask\end{tabular}} & 0.06 & 0.00 & 0.00 & 0.06 & 0.02 & 0.00 & 0.00 & 0.06 & 0.00 & 0.00 & 0.00 & 0.00 \\
    ROCKET-2 &  & 0.56 & 0.08 & 0.08 & 0.54 & 0.18 & 0.10 & 0.06 & 0.46 & 0.32 & 0.22 & 0.08 & 0.04 \\
    \textbf{\attacca{} (Ours)} &  & \XBenchBest{0.84} & \XBenchBest{0.78} & \XBenchBest{0.54} & \XBenchBest{0.66} & \XBenchBest{0.40} & \XBenchBest{0.34} & \XBenchBest{0.30} & \XBenchBest{0.74} & \XBenchBest{0.64} & \XBenchBest{0.56} & \XBenchBest{0.32} & \XBenchBest{0.28} \\
    \bottomrule
  \end{tabular}}
  \vspace{-0.1in}
\end{table}

We next extend our evaluation to long-horizon tasks, where multiple goals must be completed in sequence and each task starts from the state left by the previous one. \attacca{} completes 14--27 of 50 episodes per chain, while no baseline exceeds 4~(\cref{tab:chains}). ROCKET-2 with a current-view goal, for example, clears the first stage in up to 20\% of the episodes but almost never continues, since after a goal switch the new target is often out of view and no goal can be constructed. The gap grows beyond the first stage, suggesting that \attacca{}'s advantage accumulates across goal switches: it achieves higher next-stage completion. The largest gap occurs after mining the oak log, where \attacca{} proceeds to the diamond ores in nearly all successful episodes, compared with only one in seven for fine-tuned ROCKET-2~(\cref{fig:dpx-diamond}). \attacca{} maintains a high and stable next-stage completion rate throughout each chain, reaching 85--93\% at most transitions. Overall, it completes 5--7 times as many chains as the strongest baseline, suggesting that it substantially narrows the skill hand-off gap.

\subsection{Ablation Study}\label{sec:abl-delivery}
\begin{table}[t]
  \small
  \setlength{\tabcolsep}{3.5pt}
  \renewcommand{\arraystretch}{1.2}
  \begin{minipage}[b]{0.48\linewidth}
  \centering
  \caption{Component ablation. Success with residual feedback and phase conditioning toggled separately ($\pm$ standard error, $n{=}50$).}
  \label{tab:component-ablation}
  \vspace{-0.1in}
  \begin{tabular}{@{}cc ccc@{}}
    \toprule
    \textbf{Residual} & \textbf{Phase} & \XBenchIcon{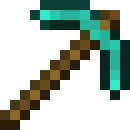} & \XBenchIcon{wolf_tamed} & \XBenchIcon{nether_portal_lit} \\
    \midrule
    -- & -- & 0.50\,{\scriptsize$\pm$.07} & 0.12\,{\scriptsize$\pm$.05} & 0.18\,{\scriptsize$\pm$.05} \\
    \checkmark & -- & \textbf{0.54}\,{\scriptsize$\pm$.07} & 0.14\,{\scriptsize$\pm$.05} & 0.08\,{\scriptsize$\pm$.04} \\
    -- & \checkmark & 0.42\,{\scriptsize$\pm$.07} & 0.28\,{\scriptsize$\pm$.06} & 0.06\,{\scriptsize$\pm$.03} \\
    \checkmark & \checkmark & \textbf{0.54}\,{\scriptsize$\pm$.07} & \textbf{0.30}\,{\scriptsize$\pm$.06} & \textbf{0.28}\,{\scriptsize$\pm$.06} \\
    \bottomrule
  \end{tabular}
  \end{minipage}\hfill
  \begin{minipage}[b]{0.48\linewidth}
  \centering
  \caption{Feedback-pathway ablation. Success with each pathway from the predicted mask to the policy ($\pm$ standard error, $n{=}50$).}
  \label{tab:pathway-ablation}
  \vspace{-0.1in}
  \begin{tabular}{@{}l ccc@{}}
    \toprule
    \textbf{Feedback pathway} & \XBenchIcon{diamond_pickaxe} & \XBenchIcon{wolf_tamed} & \XBenchIcon{nether_portal_lit} \\
    \midrule
    Cross-attention & 0.36\,{\scriptsize$\pm$.07} & 0.08\,{\scriptsize$\pm$.04} & 0.18\,{\scriptsize$\pm$.05} \\
    Layerwise FiLM & 0.42\,{\scriptsize$\pm$.07} & 0.10\,{\scriptsize$\pm$.04} & 0.18\,{\scriptsize$\pm$.05} \\
    ControlNet & 0.16\,{\scriptsize$\pm$.05} & 0.00\,{\scriptsize$\pm$.00} & 0.08\,{\scriptsize$\pm$.04} \\
    Residual (\textbf{Ours}) & \textbf{0.54}\,{\scriptsize$\pm$.07} & \textbf{0.30}\,{\scriptsize$\pm$.06} & \textbf{0.28}\,{\scriptsize$\pm$.06} \\
    \bottomrule
  \end{tabular}
  \end{minipage}
  \vspace{-0.1in}
\end{table}

\paragraph{Components.}
\cref{tab:component-ablation} ablates the core components of \attacca{} on the three long-horizon chains. We begin with dense target grounding alone, where the mask head is supervised but its prediction is not fed back to the policy, and independently add residual mask feedback~(\cref{sec:grounding}) and behavioral-phase conditioning~(\cref{sec:phase}). Neither component helps consistently on its own. Residual feedback alone barely changes Diamond Pickaxe and Wolf Feeding and lowers Nether Portal from 0.18 to 0.08, while phase conditioning alone raises Wolf Feeding from 0.12 to 0.28 but lowers the other two chains. Combining both improves all three chains by 0.04, 0.18, and 0.10 over the first row. This suggests that phase conditioning indicates when to transition from search to approach, while residual feedback provides the target location needed to act on that transition.

\paragraph{Feedback pathway.}
With target supervision and phase conditioning fixed, we compare four mask-injection pathways, all using the same mask head and supervision.
The residual pathway adds a single grounding vector to the readout token before the causal Transformer. Cross-attention lets the readout hidden state after the first layer attend to the $14\times14$ patch features, with attention biased toward the predicted mask. Layerwise FiLM modulates all tokens before every layer and ControlNet runs a trainable Transformer copy in parallel with the frozen original. 
As shown in \cref{tab:pathway-ablation}, the residual pathway performs best on all three chains. We hypothesize that a single residual input lets the policy absorb the large variation in mask presence across behavioral phases. Layerwise FiLM and ControlNet instead inject these variations into the Transformer computation, which may make the policy overly sensitive to phase-dependent changes in mask presence and scale. Cross-attention, which gives the readout direct access to mask-weighted patch features, also remains below the residual, suggesting that the target's presence and location, summarized in a single vector at the input, are sufficient for acting on the mask prediction, and that finer patch-level features add no benefit.

\subsection{Qualitative Analysis}\label{sec:exemplar-execution}
\begin{figure}[t]
  \centering
  \vspace{-0.05in}
  \begin{minipage}[t]{0.4707\linewidth}
    \centering
    \includegraphics[width=\linewidth]{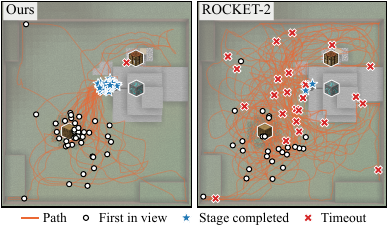}
    \vspace{-1.3em}
    \subcaption{Trajectories}\label{fig:dpx-diamond-paths}
  \end{minipage}\hfill
  \begin{minipage}[t]{0.5147\linewidth}
    \centering
    \includegraphics[width=\linewidth]{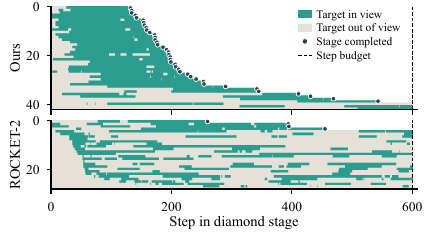}
    \vspace{-1.3em}
    \subcaption{Target visibility over time}\label{fig:dpx-diamond-raster}
  \end{minipage}
  \vspace{-0.7em}
  \caption{Diamond stage of the Diamond Pickaxe chain for all episodes that reach it (\attacca{} 42, ROCKET-2 28). (a) Trajectories. (b) Whether the diamond ore stays in view after it first appears.}
  \label{fig:dpx-diamond}
  \vspace{-0.05in}
\end{figure}

\begin{figure}[t]
\centering
\includegraphics[width=\textwidth]{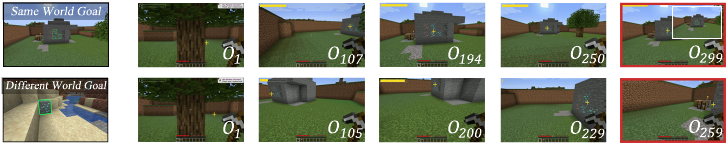}
\vspace{-0.2in}
\caption{Fine-tuned ROCKET-2 in the diamond scenario with a goal image taken far from the target in the execution world (top) or in a different world (bottom). Frames overlay its predicted target point and visibility, and red borders mark the final frame of a failed episode.}
\label{fig:failure-rocket2}
\vspace{-0.15in}
\end{figure}

\begin{wraptable}[8]{r}{0.54\textwidth}
\centering
\vspace{-0.15in}
\caption{Success by goal image, pooled over both scenarios ($\pm$ standard error, $n{=}100$).}
\label{tab:pooled-2model-400}
\vspace{-0.05in}
\footnotesize
\setlength{\tabcolsep}{2.5pt}
\renewcommand{\arraystretch}{1.25}
\resizebox{\linewidth}{!}{%
\begin{tabular}{@{}lccc@{}}
\toprule
\textbf{Policy} & Same world (close) & Same world (far) & Different world \\
\midrule
ROCKET-2 & 0.75 {\scriptsize$\pm$0.04} & 0.24 {\scriptsize$\pm$0.04} & 0.30 {\scriptsize$\pm$0.05} \\
\textbf{Ours} & \XBenchBest{0.89} {\scriptsize$\pm$0.03} & \XBenchBest{0.89} {\scriptsize$\pm$0.03} & \XBenchBest{0.81} {\scriptsize$\pm$0.04} \\
\bottomrule
\end{tabular}}
\vspace{-0.05in}
\end{wraptable}

We first examine every episode that reaches the diamond stage of the Diamond Pickaxe chain in \cref{tab:chains}, comparing \attacca{} with the fine-tuned ROCKET-2~(\cref{fig:dpx-diamond}). Once the diamond ore enters \attacca{}'s view, it stays in view almost continuously until the stage is completed~(\cref{fig:dpx-diamond-raster}), and \attacca{}'s paths converge on it. The diamond ore also enters ROCKET-2's view in 27 of its 28 episodes, but it repeatedly leaves the view afterward, and ROCKET-2's paths spread over the arena until the 600-step budget runs out. It completes the stage in only 4 of these 27 episodes, against 39 of 42 for \attacca{}, which indicates weaker target-directed control after the target is found rather than a failure to find it.
To see why, \cref{fig:failure-rocket2} follows ROCKET-2 in single rollouts of the diamond scenario with two fixed goal images, and \cref{tab:pooled-2model-400} reports success by goal image. \cref{app:failure-analysis} adds the other methods and the oak scenario.

We deliberately capture the same-world goal in the top row of \cref{fig:failure-rocket2} far from the target, so that reproducing its view and reaching the target lead to different places. Grounding works, and the predicted point is correct at $o_{107}$, but after nearing the target at $o_{194}$ the agent backs away and settles at $o_{299}$, where its view reproduces the goal view, instead of interacting. With the different-world goal in the bottom row, grounding starts out correct at $o_{200}$, but visibility collapses at $o_{229}$ and the agent walks past the diamond, as in \cref{fig:dpx-diamond}. This suggests that ROCKET-2, trained with goals from the same world as its demonstrations, acts by reproducing the goal view rather than by pursuing the masked target. Reproducing the goal view works only for close same-world goals, whose view already places the agent within reach. With these goals ROCKET-2 reaches 0.75 success, but only 0.24 with far goals and 0.30 with different-world goals, whereas \attacca{} stays at 0.81--0.89 for all three~(\cref{tab:pooled-2model-400}). \attacca{} is trained with goal images from other worlds, so reproducing the goal view cannot lead it to the target, and it must pursue the masked target instead. Given the goal image that ROCKET-2 receives in the bottom row of \cref{fig:failure-rocket2}, \attacca{} switches from Search to Approach when its predicted mask first activates and mines all three diamond ores~(\cref{app:failure-analysis}).

\section{Conclusion}\label{sec:conclusion}
We presented \attacca{}, a visual goal-conditioned policy for goal-directed embodied control under state continuity, in which each task starts from the state left by the previous one and the goal image comes from a different world. Context-decoupled goal sampling, goal-conditioned target grounding, and behavioral-phase conditioning let a single policy search for, approach, and interact with targets from such inherited states. \attacca{} achieves the highest success on all short-horizon tasks for both seen and held-out classes and completes 5--7 times as many long-horizon chains as the strongest baseline. Extending \attacca{} to other embodied domains and to learned planners is left for future work, and \cref{app:limitations} discusses the limitations of our evaluation.

\subsection*{AI use statement}
In this work, we used generative AI tools to aid in correcting grammatical errors and improving wording. We have not used generative AI tools for data collection, research conceptualization, experiments, or analysis. We have reviewed all AI-assisted text, and we take responsibility for the final content of this work, including text, claims or artifacts produced with the aid of generative AI.

\ifnum\referencepreview=0
\bibliography{references}
\bibliographystyle{iclr2027_conference}
\fi

\appendix
\clearpage
\raggedbottom
\section{Example of Context-Decoupled Goal Sampling}\label{app:sampling}
\begin{figure}[H]
\centering
\includegraphics[width=\textwidth]{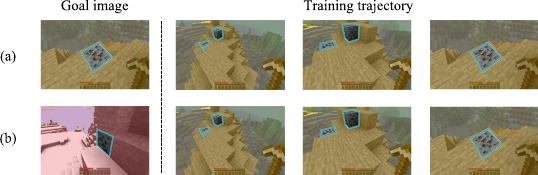}
\caption{(a) Context-correlated sampling draws goal images from the same world as the training trajectory. (b) Our context-decoupled sampling draws goal images of the same target class from a different world.}
\label{fig:sampling}
\end{figure}

Figure~\ref{fig:sampling} pairs one training trajectory with two goal images. Under context-correlated sampling (a), the goal image comes from the same world as the trajectory and shows the same surrounding terrain, so matching the scene can substitute for locating the target. Under context-decoupled sampling (b), the goal image shows the same target class in another world with different terrain. It shares no scene content with the observations, and the appearance of the target is the only usable cue.

\begin{figure}[H]
\centering
\includegraphics[width=\textwidth]{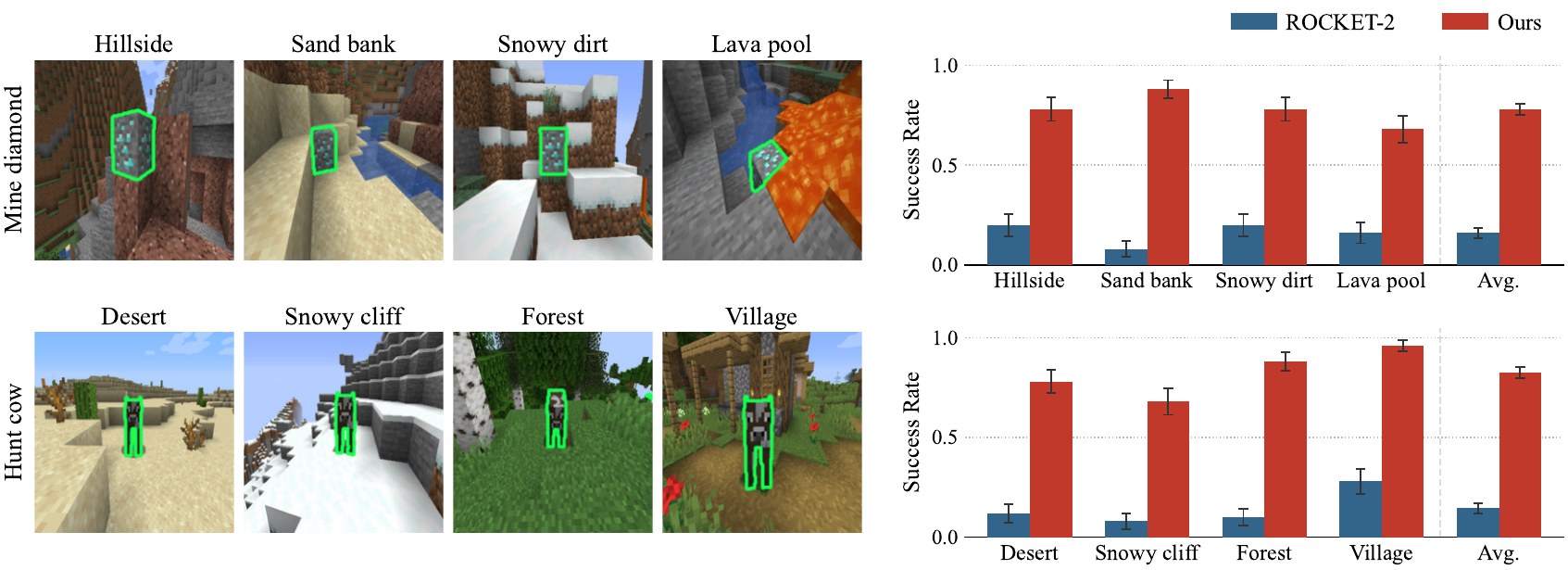}
\caption{Success with goal images of four different backgrounds, 50 episodes per goal.}
\label{fig:goal-background}
\end{figure}

\paragraph{Robustness to the goal background.}
We evaluate \attacca{} and the fine-tuned ROCKET-2 on the mine diamond and hunt cow tasks with four goal images of different backgrounds, over 50 episodes per goal~(\cref{fig:goal-background}). \attacca{} succeeds in 0.68--0.96 of the episodes for every goal image, whereas ROCKET-2 stays at 0.08--0.28, so the gain of \attacca{} holds for every goal background rather than depending on a particular goal image.

\section{Search-to-Interact Dataset}\label{app:dataset}
\begin{figure}[H]
  \centering
  \setlength{\tabcolsep}{1.5pt}
  \renewcommand{\arraystretch}{0}
  \begin{tabular}{cccc}
    \includegraphics[width=0.24\textwidth]{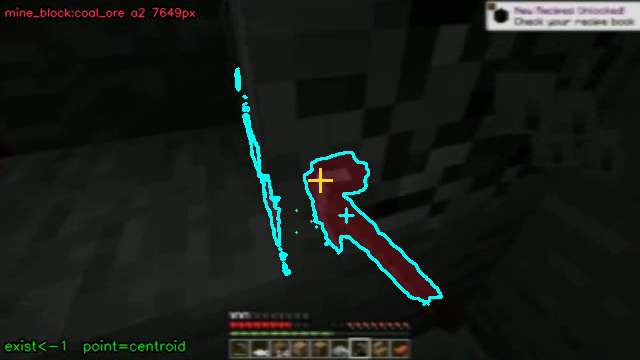} &
    \includegraphics[width=0.24\textwidth]{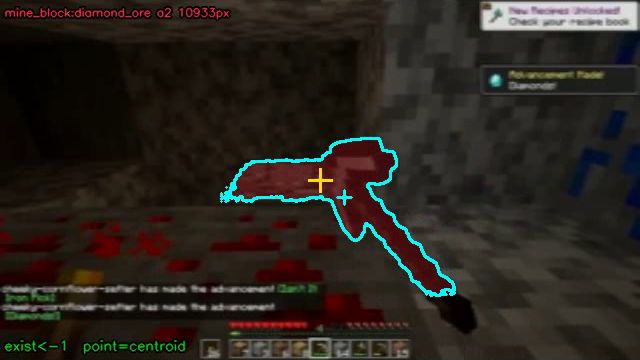} &
    \includegraphics[width=0.24\textwidth]{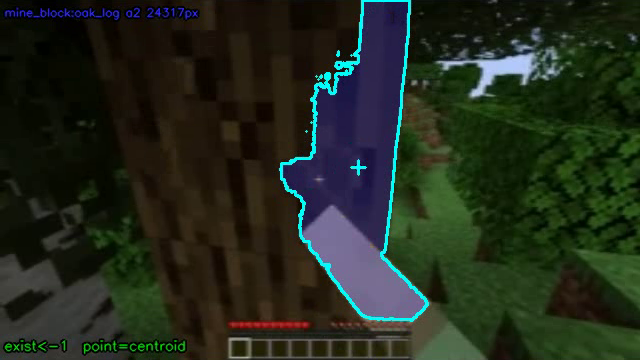} &
    \includegraphics[width=0.24\textwidth]{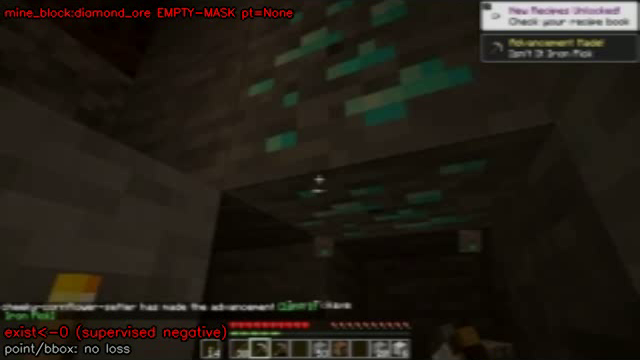} \\
  \end{tabular}
  \caption{Errors of hindsight labels. In the two left frames, the center prompt selects the held tool, which is then labeled as \emph{coal ore} and \emph{diamond ore} while the ore in view stays unlabeled. In the third frame, the oak-log mask spreads onto the agent's body and shifts the mask centroid. In the right frame, diamond ore fills the view, but the frame is labeled as target-absent.}
  \label{fig:label-failures}
\end{figure}

\paragraph{Limitations of hindsight labeling.}
Prior goal-conditioned agents are trained with labels produced in hindsight~\papercite{cai2025rocket2}. An interaction event is detected in gameplay, a segmentation model is prompted at the center of the frame just before the event, and the resulting mask is tracked backward in time. Figure~\ref{fig:label-failures} shows typical errors of this process, including masks on the held tool, masks that spread onto the player's arm, and empty masks while the target is still visible. Visibility and grounding supervision from such labels is noisy, and search frames cannot be identified reliably.

\begin{figure}[H]
\centering
\includegraphics[width=\textwidth]{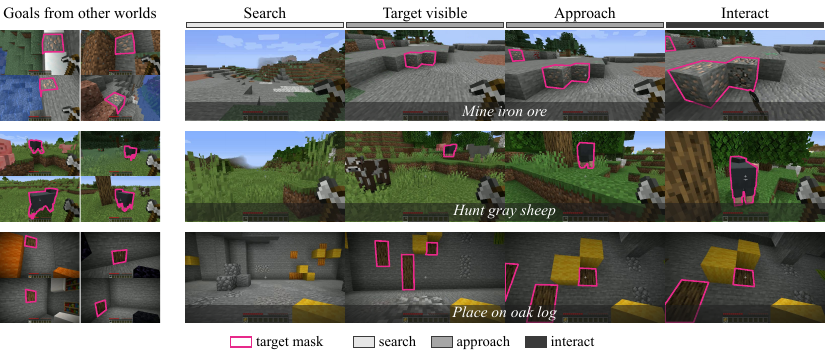}
\caption{Search-to-interact demonstrations. For each task, one human demonstration is shown at four moments, from the start of the episode through the first view of the target and the approach to the interaction, with its target mask (outline) and behavioral phase (bar above each frame). On the left are example goal images of the same target class, each taken from a demonstration recorded in a different world and shown with its target mask.}
\label{fig:dataset}
\end{figure}

\paragraph{The HumanPlay dataset.}
We collect 1,160 demonstrations with 284,961 frames in total, recorded by experienced Minecraft players, each in a different world. A demonstration starts at a random spawn point in a new world and covers the full progression from search to approach and interaction for one of the seven training classes of its task~(Figure~\ref{fig:dataset}). The demonstrator continues until a task-specific interaction quota is reached, so a single episode contains several search-to-interact cycles. Every scene contains objects of other classes along with the target instances, and since the demonstrator interacts only with the requested target, the demonstrations also show target selection among distractors. The dataset contains Mine (822 episodes, 209,593 frames), Hunt (198 episodes, 47,991 frames), and Place (140 episodes, 27,377 frames), and these counts include the 10\% of episodes held out for validation. The environment is built on MineStudio~\papercite{cai2024minestudio}.

\paragraph{Labels.}
Per-frame target masks come from the renderer, which back-projects the depth buffer and assigns visible pixels to target instances. The mask covers all visible surfaces of the instances of the requested target class. Frames with no visible target, 35\% of all frames, receive an all-zero mask and thus supervise target absence. Each frame is also labeled with a behavioral phase. A frame is Interact if it contains an attack or use event on the selected target, Approach if the target is visible or has been seen and is temporarily out of view, and Search otherwise. Search covers 33\% of the frames, Approach 47\%, and Interact 20\%. These labels are used only as training targets and are never given to the policy at inference.

\section{Implementation Details}\label{app:recipe}
Table~\ref{tab:impl} lists the architecture and training settings of \attacca{}. The policy is initialized from the released ROCKET-2 checkpoint~\papercite{cai2025rocket2}. Goal images are sampled as described in \cref{sec:relabel}, using only target-visible frames, which are resized to $224\times224$. A new reference demonstration and goal frame are drawn each time a training sample is loaded.

$\mathcal{L}_{\mathrm{BC}}$ is summed over the 128 steps of a window and $\mathcal{L}_{\phi}$ is averaged over frames. $\mathcal{L}_{\mathrm{target}}$ is the mean of its averages over frames with and without a visible target. Varying each loss weight over a 16$\times$ range with the other held at its default changes success on Mine by at most 4.5 points for $\lambda_{\mathrm{target}}$ and 6.0 points for $\lambda_\phi$~(Table~\ref{tab:lambda-sweep}), within the $\pm$6.7-point 95\% binomial confidence interval at $n{=}200$.
\begin{table}[H]
  \centering
  \caption{Architecture and training settings of \attacca{}.}
  \label{tab:impl}
  \small
  \renewcommand{\arraystretch}{1.15}
  \begin{tabular}{@{}ll@{}}
    \toprule
    \textbf{Component} & \textbf{Setting} \\
    \midrule
    Observation and goal image & $224\times224$ RGB \\
    Visual encoder & DINO ViT-B/16~\papercite{Caron_2021_ICCV}, frozen, 196 patch tokens \\
    Goal-mask encoder & ViT-Tiny/16, ImageNet-21k initialization, trainable \\
    Goal-fusion module & Transformer encoder, 4 layers, 8 heads, 9 query tokens \\
    Causal Transformer & Transformer-XL~\papercite{dai-etal-2019-transformer}, 4 layers, width 1024 \\
    Mask head $h_{\mathrm{mask}}$ & per-patch MLP over the $14\times14$ patch grid \\
    Residual encoder inputs & mask-weighted pooled $P_t$, $p_t$, and the mean, maximum, and entropy of $p_t$ \\
    Residual projection & $1024\times1024$, zero-initialized \\
    Phase head $h_\phi$ & linear, with zero-initialized $W_\beta$ and $W_\gamma$ \\
    Action space & VPT mouse and keyboard~\papercite{NEURIPS2022_9c7008af} \\
    \midrule
    Initialization & released ROCKET-2 checkpoint~\papercite{cai2025rocket2} \\
    Training data & 1,160 episodes, 284,961 frames, 10\% held out for validation \\
    Window length & 128 frames \\
    Batch size & 12 windows (3 GPUs $\times$ 4) \\
    Optimization steps & 1,350, with 200 warmup steps \\
    Learning rate & $2\times10^{-5}$ \\
    Weight decay & $10^{-3}$ \\
    Loss weights & $\lambda_{\mathrm{target}}=0.5$, $\lambda_\phi=0.1$ \\
    Training seed & 2026 \\
    \bottomrule
  \end{tabular}
\end{table}

\begin{table}[H]
  \centering
  \caption{Loss-weight sensitivity on Mine ($n{=}200$ per
  cell). Each weight is varied with the other held at its default; bold marks the defaults used in all main results.}
  \label{tab:lambda-sweep}
  \small
  \setlength{\tabcolsep}{6pt}
  \renewcommand{\arraystretch}{1.2}
  \begin{tabular}{@{}l ccccc@{}}
    \toprule
    $\lambda_{\mathrm{target}}$ & 0.125 & 0.25 & \textbf{0.5} & 1.0 & 2.0 \\
    \cmidrule(l){2-6}
    Mine & 0.355 & 0.345 & 0.390 & 0.370 & 0.380 \\
    \midrule
    $\lambda_{\phi}$ & 0.025 & 0.05 & \textbf{0.1} & 0.2 & 0.4 \\
    \cmidrule(l){2-6}
    Mine & 0.330 & 0.355 & 0.390 & 0.375 & 0.370 \\
    \bottomrule
  \end{tabular}
\end{table}

\clearpage
\section{Short-Horizon Task Details}\label{app:benchmark}
\begin{figure}[H]
\centering
\includegraphics[width=\textwidth]{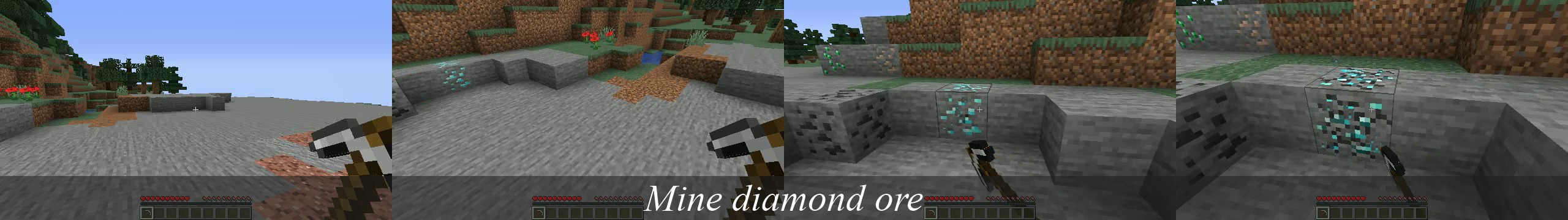}\\[2pt]
\includegraphics[width=\textwidth]{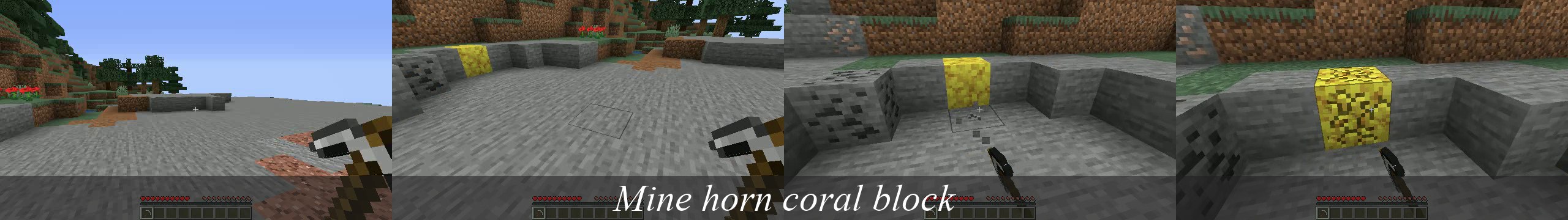}\\[2pt]
\includegraphics[width=\textwidth]{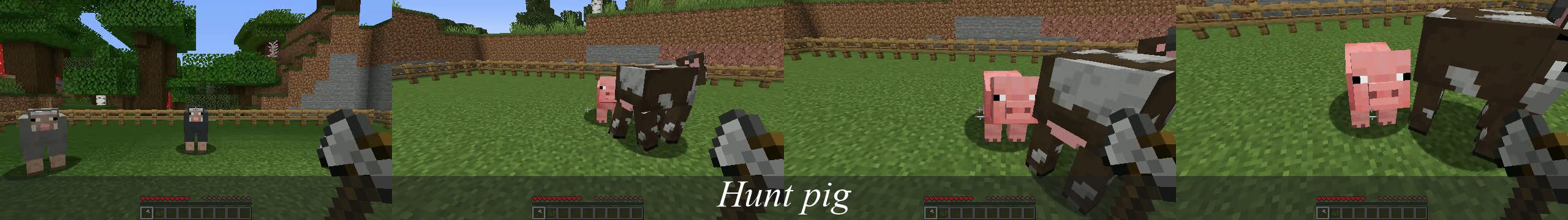}\\[2pt]
\includegraphics[width=\textwidth]{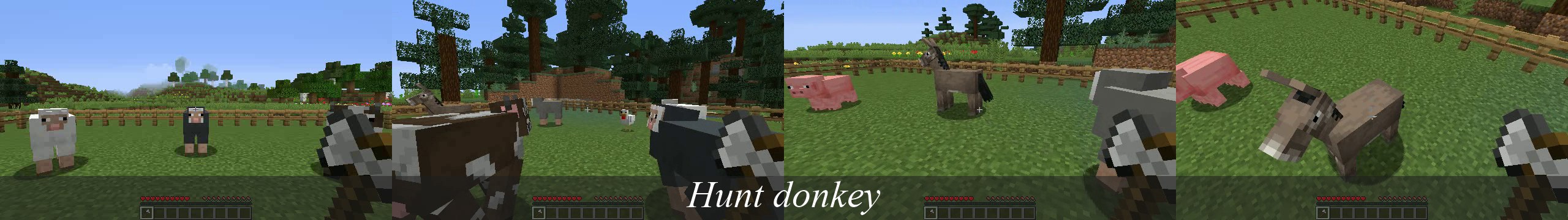}\\[2pt]
\includegraphics[width=\textwidth]{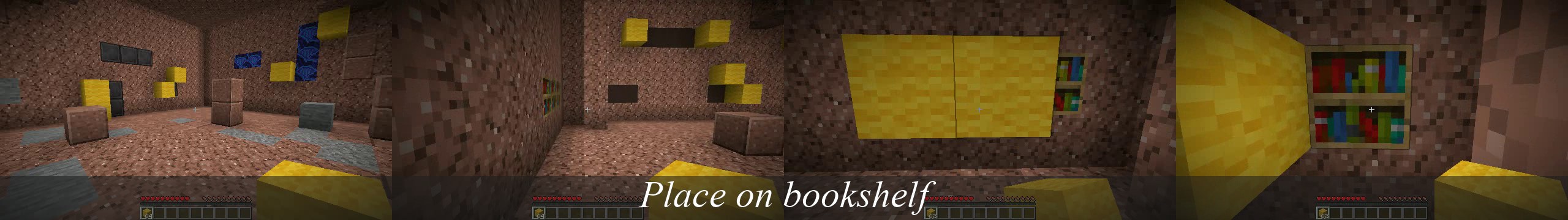}\\[2pt]
\includegraphics[width=\textwidth]{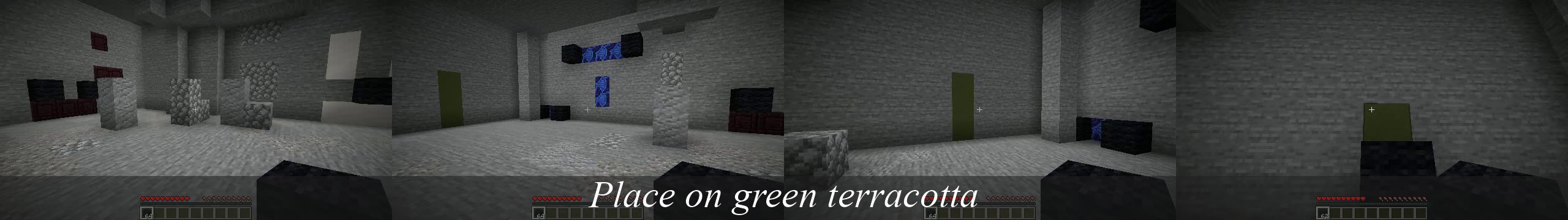}
\caption{Short-horizon tasks. Each row shows four frames of one successful \attacca{} episode, from the first frame through target discovery and approach to the interaction. For each task, the upper row shows a class seen in training and the lower row a held-out class.}
\label{fig:short-tasks}
\end{figure}

In each short-horizon task, a single target starts out of view among objects of other classes, and the goal image comes from another world (Figure~\ref{fig:short-tasks}). Each task has ten target classes~\mbox{(Tables~\ref{tab:detail-mine}--\ref{tab:detail-place})}, seven seen in training and three held out (OOD), and all methods are evaluated on the same scenes and action budgets with 200 episodes per task, 140 of them on seen classes.

\paragraph{Mine.}
Mine uses seven ores seen in training and three held-out coral blocks on common terrain, each evaluated in ten worlds with two policy seeds. The staged target is swapped between classes while the terrain stays fixed, and held-out classes never appear as distractors for other classes.

\paragraph{Hunt and Place.}
Hunt uses seven seen mobs, including four sheep colors, and three held-out mobs, each in ten layouts. Place uses seven seen and three held-out target blocks, each in ten worlds, and the agent places its held block onto the target. Outcomes and off-target interactions are scored from the world state, which the scorer can query but the policy cannot observe.

\paragraph{Clean success.}
An episode is a clean success if it achieves the requested outcome without any wrong-class interaction, which means breaking the target block without breaking blocks of other classes in Mine, killing the target without killing a bystander in Hunt, and filling every target cell without off-target placements in Place. The criterion does not change episode termination or action budgets, so an episode that reaches the goal after a wrong-class interaction counts as a failure.

\begin{table}[H]%
  \centering
  \caption{Mine, per-class clean success and interaction precision (share of an episode's interactions on the target class). Each class is tested 20 times per method.}
  \label{tab:detail-mine}
  \small
  \setlength{\tabcolsep}{3.0pt}
  \renewcommand{\arraystretch}{1.26}
  \resizebox{\textwidth}{!}{\begin{tabular}{@{}l|c|ccccccc|ccc|cc@{}}
    \toprule
    \multirow{2}{*}{\textbf{Policy}} & \multirow{2}{*}{\textbf{Goal modality}} & \multicolumn{7}{c|}{ID} & \multicolumn{3}{c|}{OOD} & \multirow{2}{*}{\textbf{avg}} & \multirow{2}{*}{\textbf{prec}} \\
    & & \XBenchIcon{coal_ore} & \XBenchIcon{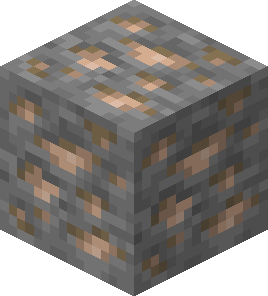} & \XBenchIcon{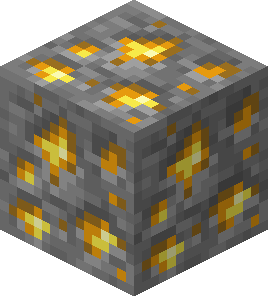} & \XBenchIcon{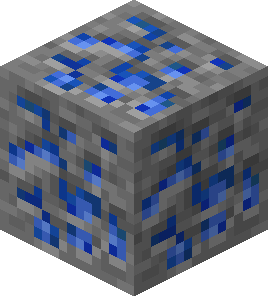} & \XBenchIcon{diamond_ore} & \XBenchIcon{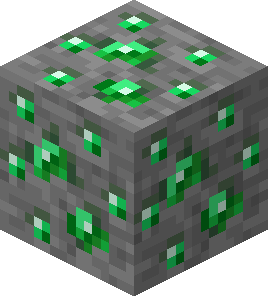} & \XBenchIcon{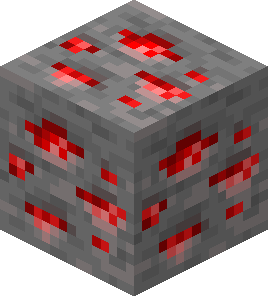} & \XBenchIcon{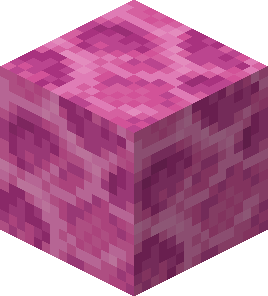} & \XBenchIcon{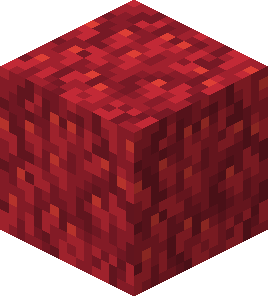} & \XBenchIcon{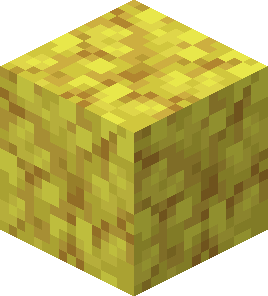} & & \\
    \midrule
    STEVE-1 & \multirow{2}{*}{Text} & 0.00 & 0.00 & 0.05 & 0.00 & 0.00 & 0.00 & 0.00 & 0.00 & 0.00 & 0.00 & 0.005 & 0.091 \\
    JARVIS-VLA &  & 0.15 & 0.00 & 0.10 & 0.15 & 0.05 & 0.15 & 0.20 & 0.00 & 0.00 & 0.00 & 0.080 & 0.727 \\
    \cmidrule(lr){1-14}
    STEVE-1 & \multirow{2}{*}{Video} & 0.05 & 0.05 & 0.05 & 0.00 & 0.05 & 0.00 & 0.00 & 0.00 & 0.00 & 0.00 & 0.020 & 0.210 \\
    GROOT &  & 0.00 & 0.00 & 0.05 & 0.00 & 0.05 & 0.00 & 0.00 & 0.00 & 0.00 & 0.00 & 0.010 & 0.118 \\
    \cmidrule(lr){1-14}
    ROCKET-1 & \multirow{2}{*}{\begin{tabular}[c]{@{}c@{}}Current view\\$+$ Molmo $+$ SAM\end{tabular}} & 0.00 & 0.00 & 0.00 & 0.05 & 0.05 & 0.00 & 0.05 & 0.00 & 0.05 & 0.00 & 0.020 & 0.237 \\
    ROCKET-2 &  & 0.00 & 0.05 & 0.15 & 0.10 & 0.05 & 0.00 & 0.05 & 0.00 & 0.10 & 0.00 & 0.050 & 0.580 \\
    \cmidrule(lr){1-14}
    ROCKET-2$^{\dagger}$ & \multirow{3}{*}{\begin{tabular}[c]{@{}c@{}}Different world image\\$+$ mask\end{tabular}} & 0.00 & 0.00 & 0.00 & 0.05 & 0.00 & 0.00 & 0.00 & 0.00 & 0.00 & 0.00 & 0.005 & 0.333 \\
    ROCKET-2 &  & 0.25 & 0.40 & 0.15 & 0.35 & 0.20 & 0.25 & 0.25 & 0.10 & 0.20 & \XBenchBest{0.20} & 0.235 & 0.739 \\
    \textbf{Ours} &  & \XBenchBest{0.45} & \XBenchBest{0.45} & \XBenchBest{0.20} & \XBenchBest{0.60} & \XBenchBest{0.45} & \XBenchBest{0.30} & \XBenchBest{0.70} & \XBenchBest{0.30} & \XBenchBest{0.25} & \XBenchBest{0.20} & \XBenchBest{0.390} & \XBenchBest{0.778} \\
    \bottomrule
  \end{tabular}}
\end{table}

\begin{table}[H]
  \centering
  \caption{Hunt, per-class clean success and interaction precision (share of an episode's interactions on the target class). Each class is tested 20 times per method.}
  \label{tab:detail-hunt}
  \small
  \setlength{\tabcolsep}{3.0pt}
  \renewcommand{\arraystretch}{1.26}
  \resizebox{\textwidth}{!}{\begin{tabular}{@{}l|c|ccccccc|ccc|cc@{}}
    \toprule
    \multirow{2}{*}{\textbf{Policy}} & \multirow{2}{*}{\textbf{Goal modality}} & \multicolumn{7}{c|}{ID} & \multicolumn{3}{c|}{OOD} & \multirow{2}{*}{\textbf{avg}} & \multirow{2}{*}{\textbf{prec}} \\
    & & \XBenchIcon{cow} & \XBenchIcon{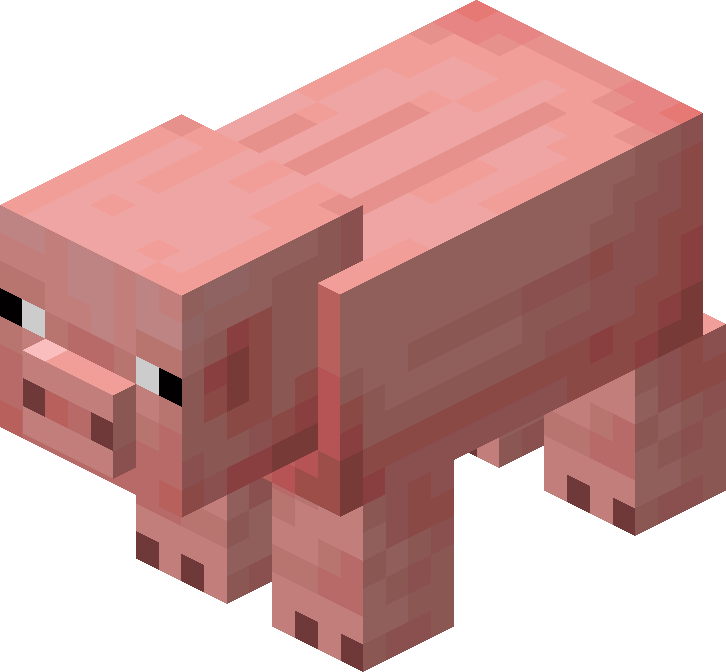} & \XBenchIcon{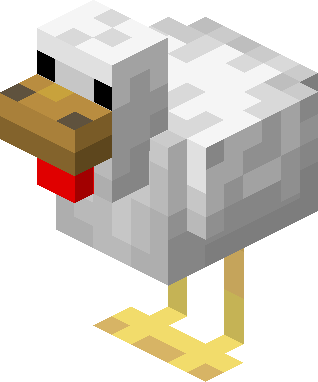} & \XBenchIcon{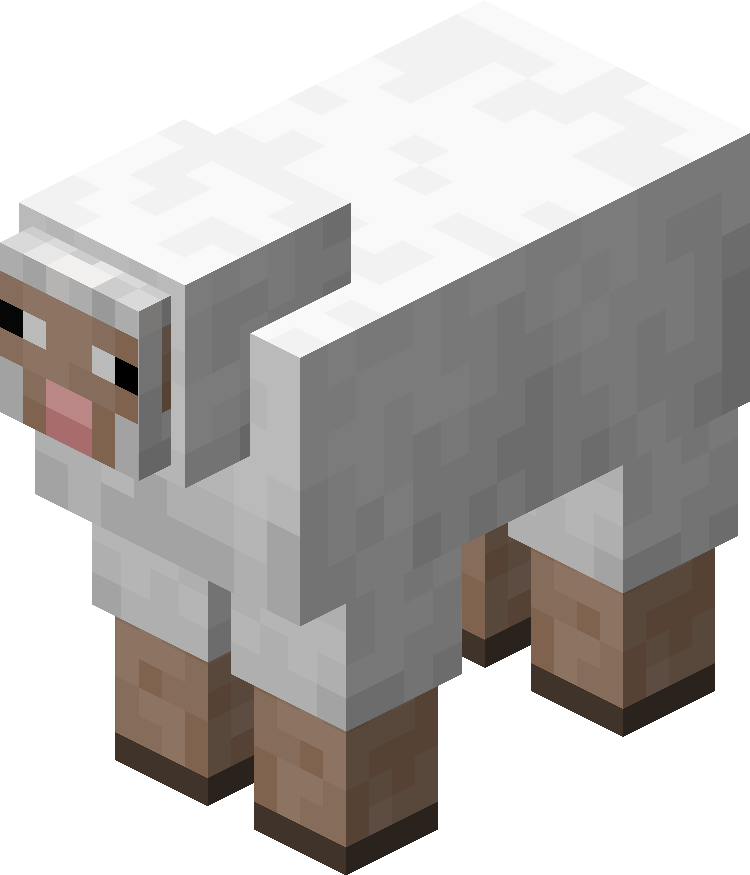} & \XBenchIcon{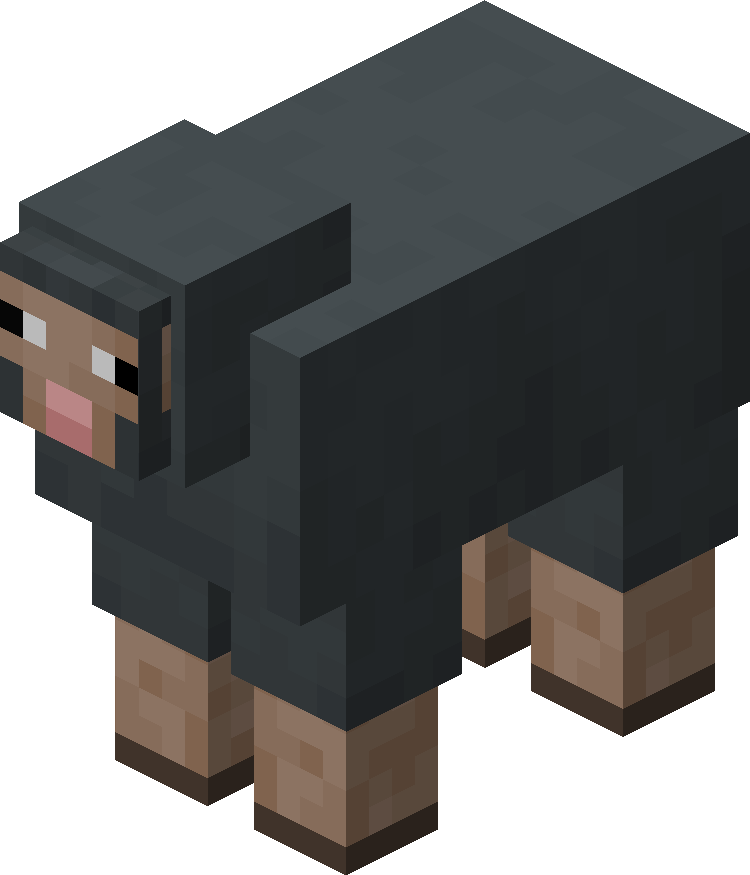} & \XBenchIcon{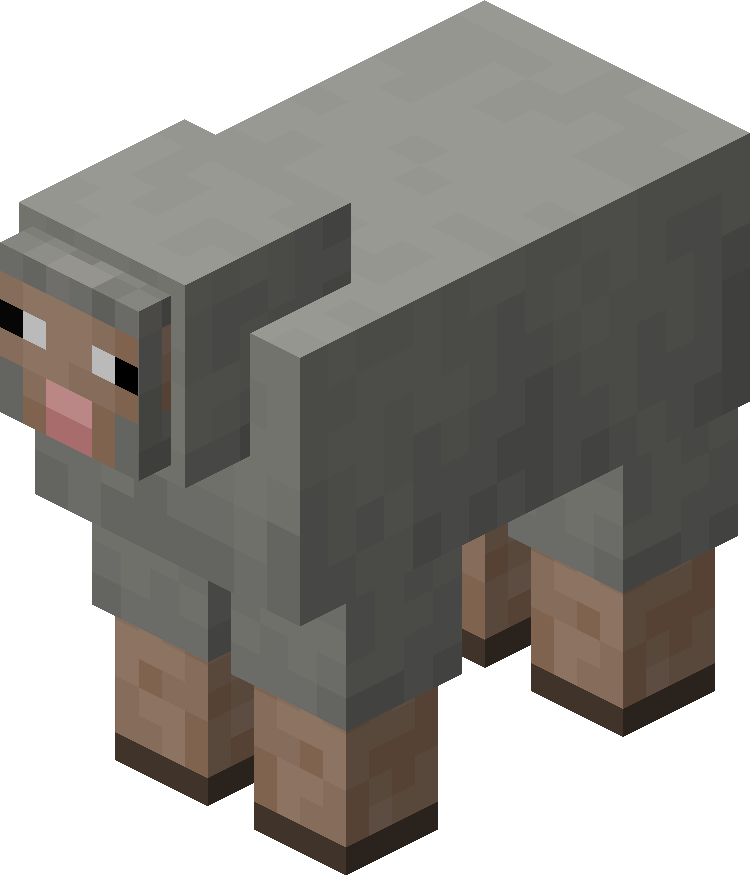} & \XBenchIcon{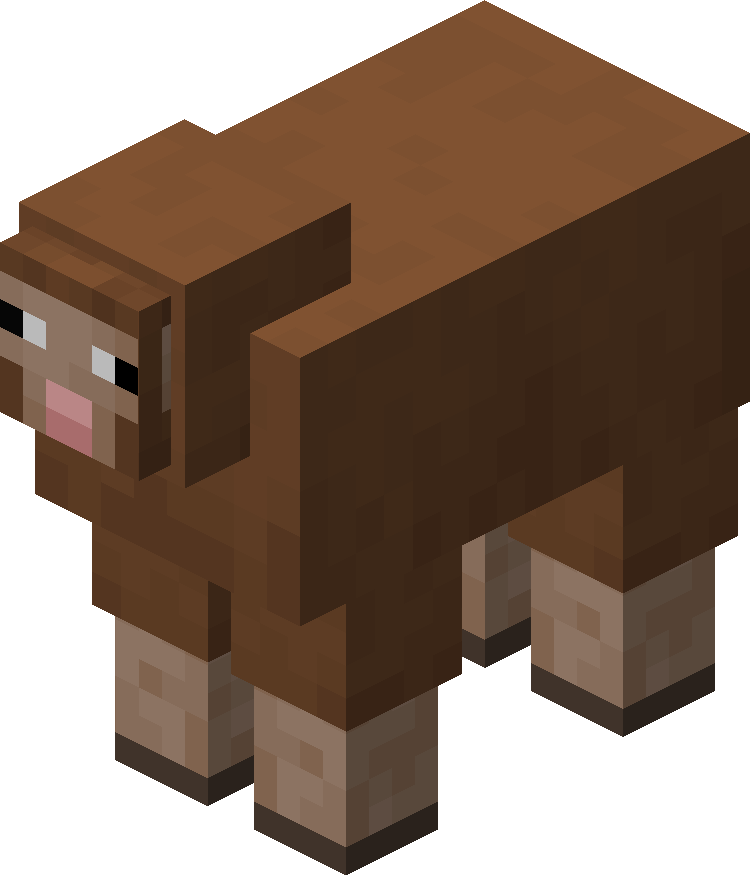} & \XBenchIcon{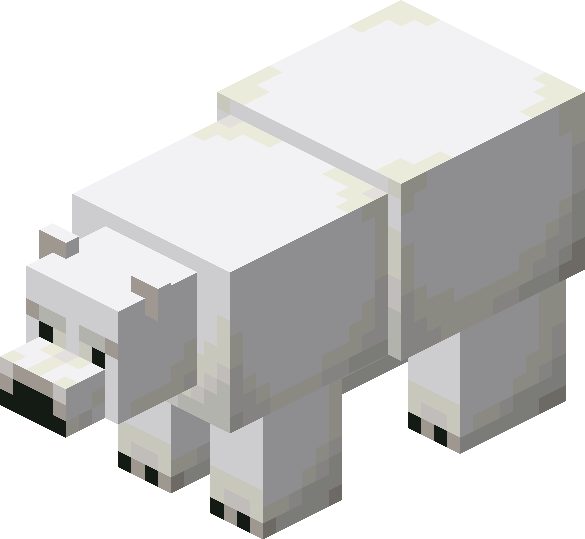} & \XBenchIcon{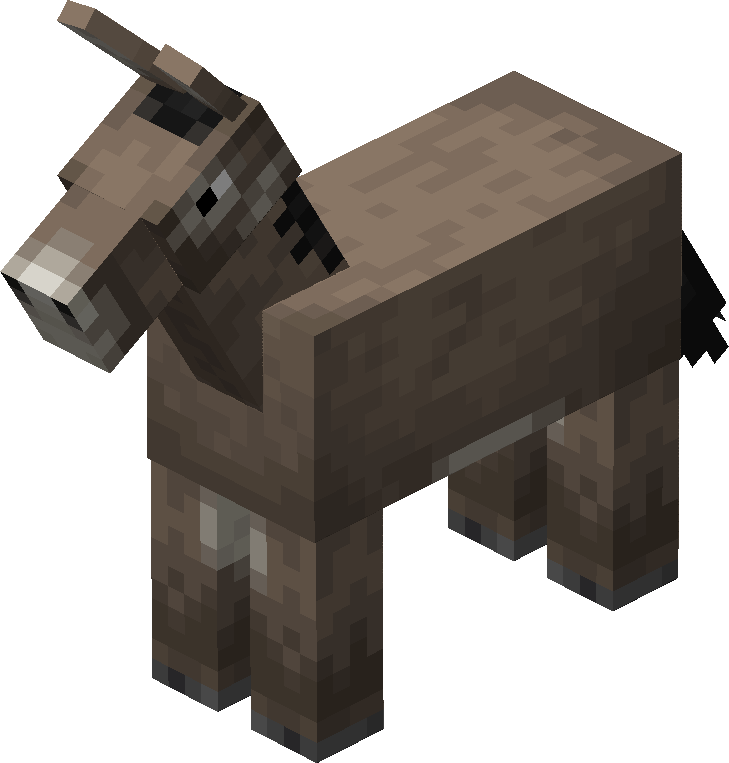} & \XBenchIcon{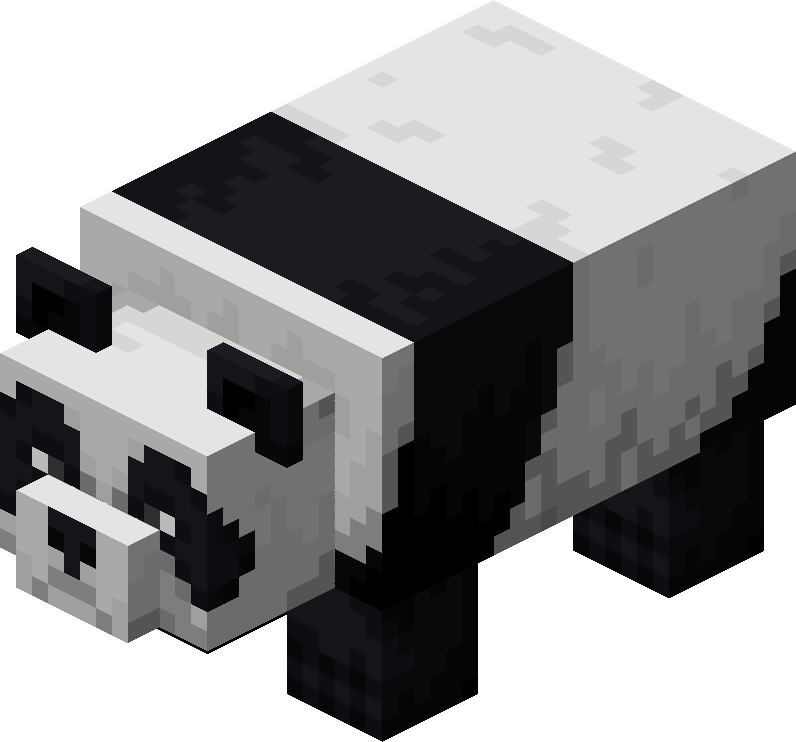} & & \\
    \midrule
    STEVE-1 & \multirow{2}{*}{Text} & 0.00 & 0.00 & 0.05 & 0.30 & 0.15 & 0.05 & 0.00 & 0.00 & 0.00 & 0.00 & 0.055 & 0.180 \\
    JARVIS-VLA &  & 0.30 & 0.25 & 0.15 & 0.20 & 0.25 & 0.00 & 0.25 & 0.00 & 0.05 & 0.20 & 0.165 & 0.468 \\
    \cmidrule(lr){1-14}
    STEVE-1 & \multirow{2}{*}{Video} & 0.00 & 0.30 & 0.00 & 0.05 & 0.05 & 0.00 & 0.00 & 0.00 & 0.00 & 0.00 & 0.040 & 0.158 \\
    GROOT &  & 0.15 & 0.00 & 0.00 & 0.15 & 0.00 & 0.05 & 0.00 & 0.05 & 0.00 & 0.00 & 0.040 & 0.127 \\
    \cmidrule(lr){1-14}
    ROCKET-1 & \multirow{2}{*}{\begin{tabular}[c]{@{}c@{}}Current view\\$+$ Molmo $+$ SAM\end{tabular}} & 0.00 & 0.30 & \XBenchBest{0.35} & 0.15 & 0.05 & 0.15 & 0.30 & 0.10 & 0.20 & 0.20 & 0.180 & 0.324 \\
    ROCKET-2 &  & 0.00 & \XBenchBest{0.45} & \XBenchBest{0.35} & 0.15 & 0.00 & 0.00 & 0.15 & \XBenchBest{0.30} & 0.35 & 0.25 & 0.200 & 0.443 \\
    \cmidrule(lr){1-14}
    ROCKET-2$^{\dagger}$ & \multirow{3}{*}{\begin{tabular}[c]{@{}c@{}}Different world image\\$+$ mask\end{tabular}} & 0.30 & 0.05 & 0.00 & 0.05 & 0.15 & 0.05 & 0.05 & 0.15 & 0.05 & 0.15 & 0.100 & 0.232 \\
    ROCKET-2 &  & 0.25 & 0.30 & 0.10 & 0.15 & 0.30 & 0.15 & 0.10 & 0.10 & 0.15 & 0.25 & 0.185 & 0.558 \\
    \textbf{Ours} &  & \XBenchBest{0.75} & 0.20 & \XBenchBest{0.35} & \XBenchBest{0.50} & \XBenchBest{0.45} & \XBenchBest{0.60} & \XBenchBest{0.50} & 0.15 & \XBenchBest{0.45} & \XBenchBest{0.80} & \XBenchBest{0.475} & \XBenchBest{0.725} \\
    \bottomrule
  \end{tabular}}
\end{table}

\begin{table}[H]
  \centering
  \caption{Place, per-class clean success and interaction precision (share of an episode's interactions on the target class). Each class is tested 20 times per method.}
  \label{tab:detail-place}
  \small
  \setlength{\tabcolsep}{3.0pt}
  \renewcommand{\arraystretch}{1.26}
  \resizebox{\textwidth}{!}{\begin{tabular}{@{}l|c|ccccccc|ccc|cc@{}}
    \toprule
    \multirow{2}{*}{\textbf{Policy}} & \multirow{2}{*}{\textbf{Goal modality}} & \multicolumn{7}{c|}{ID} & \multicolumn{3}{c|}{OOD} & \multirow{2}{*}{\textbf{avg}} & \multirow{2}{*}{\textbf{prec}} \\
    & & \XBenchIcon{oak_log} & \XBenchIcon{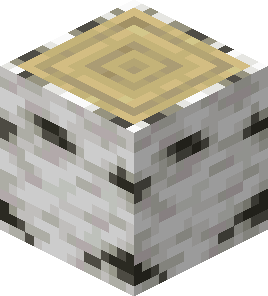} & \XBenchIcon{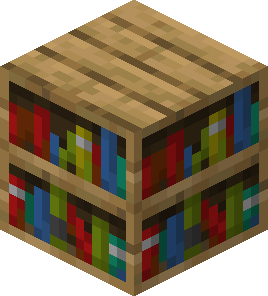} & \XBenchIcon{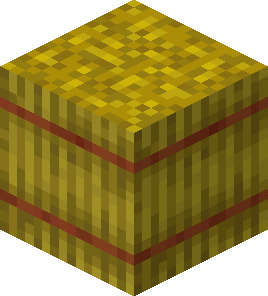} & \XBenchIcon{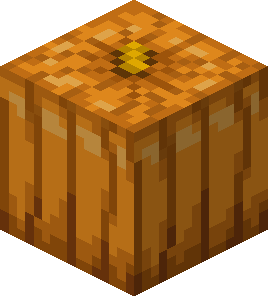} & \XBenchIcon{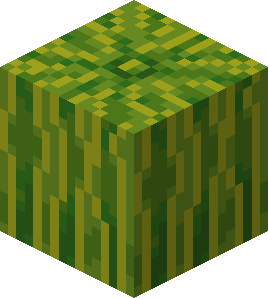} & \XBenchIcon{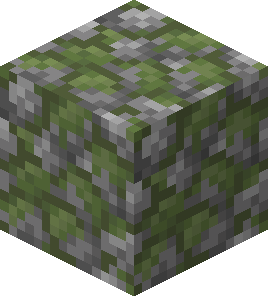} & \XBenchIcon{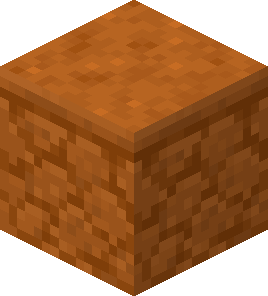} & \XBenchIcon{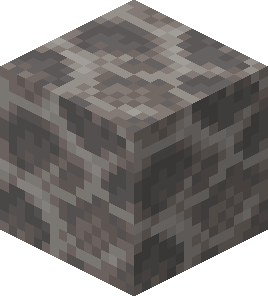} & \XBenchIcon{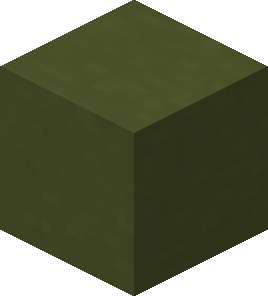} & & \\
    \midrule
    STEVE-1 & \multirow{2}{*}{Text} & 0.00 & 0.00 & 0.00 & 0.00 & 0.00 & 0.00 & 0.00 & 0.00 & 0.00 & 0.00 & 0.000 & 0.042 \\
    JARVIS-VLA &  & 0.00 & 0.05 & 0.25 & 0.10 & 0.20 & 0.15 & 0.10 & 0.00 & 0.00 & 0.00 & 0.085 & 0.521 \\
    \cmidrule(lr){1-14}
    STEVE-1 & \multirow{2}{*}{Video} & 0.00 & 0.00 & 0.00 & 0.00 & 0.00 & 0.00 & 0.00 & 0.00 & 0.00 & 0.00 & 0.000 & 0.000 \\
    GROOT &  & 0.00 & 0.00 & 0.00 & 0.00 & 0.00 & 0.00 & 0.00 & 0.00 & 0.00 & 0.00 & 0.000 & 0.002 \\
    \cmidrule(lr){1-14}
    ROCKET-1 & \multirow{2}{*}{\begin{tabular}[c]{@{}c@{}}Current view\\$+$ Molmo $+$ SAM\end{tabular}} & 0.05 & 0.05 & 0.10 & 0.15 & 0.00 & 0.05 & 0.00 & 0.05 & 0.00 & 0.00 & 0.045 & 0.081 \\
    ROCKET-2 &  & 0.00 & 0.10 & 0.05 & 0.00 & 0.00 & 0.00 & 0.00 & 0.00 & 0.00 & 0.00 & 0.015 & 0.035 \\
    \cmidrule(lr){1-14}
    ROCKET-2$^{\dagger}$ & \multirow{3}{*}{\begin{tabular}[c]{@{}c@{}}Different world image\\$+$ mask\end{tabular}} & 0.00 & 0.00 & 0.00 & 0.00 & 0.00 & 0.00 & 0.00 & 0.00 & 0.00 & 0.00 & 0.000 & 0.216 \\
    ROCKET-2 &  & 0.10 & 0.35 & 0.40 & 0.35 & 0.35 & 0.35 & 0.10 & 0.15 & 0.05 & 0.05 & 0.225 & 0.567 \\
    \textbf{Ours} &  & \XBenchBest{0.35} & \XBenchBest{0.70} & \XBenchBest{0.55} & \XBenchBest{0.65} & \XBenchBest{0.65} & \XBenchBest{0.60} & \XBenchBest{0.35} & \XBenchBest{0.40} & \XBenchBest{0.35} & \XBenchBest{0.10} & \XBenchBest{0.470} & \XBenchBest{0.579} \\
    \bottomrule
  \end{tabular}}
\end{table}

\section{Long-Horizon Task Details}\label{app:helpers}
\begin{figure}[H]
\centering
\includegraphics[width=\textwidth]{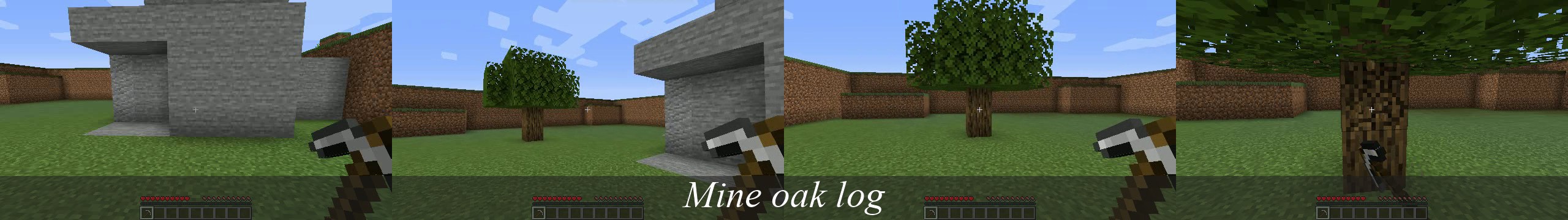}\\[2pt]
\includegraphics[width=\textwidth]{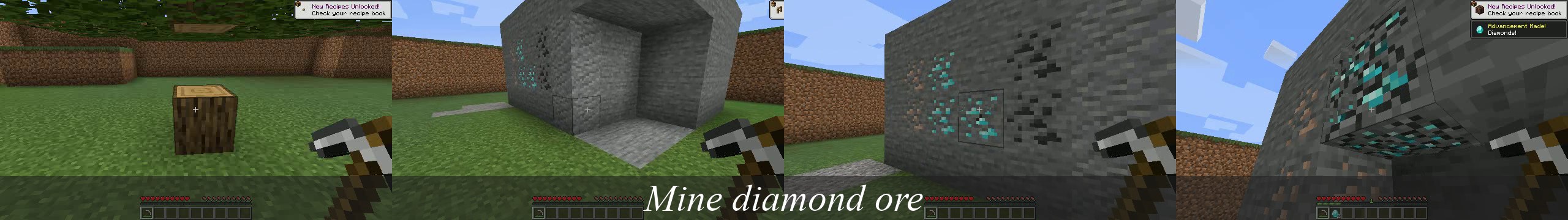}\\[2pt]
\includegraphics[width=\textwidth]{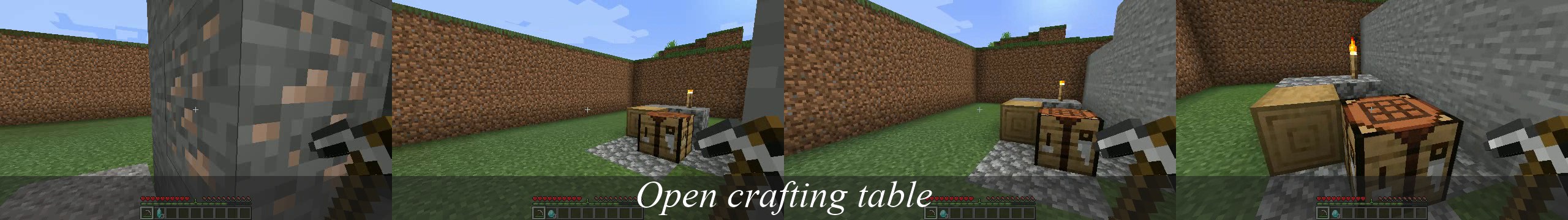}
\caption{Diamond Pickaxe chain (oak log$\to$diamond ore$\to$crafting table). Each row is one stage of a single successful \attacca{} episode, shown from the first frame of the stage through target discovery and approach to the interaction.}
\label{fig:chain-dpx}
\end{figure}

\begin{figure}[H]
\centering
\includegraphics[width=\textwidth]{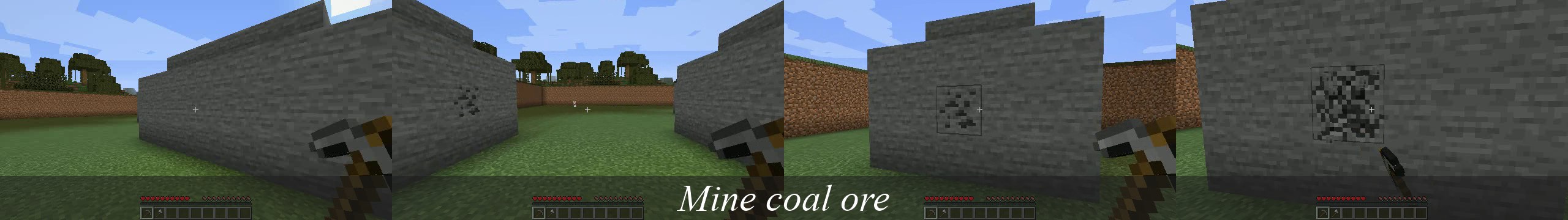}\\[2pt]
\includegraphics[width=\textwidth]{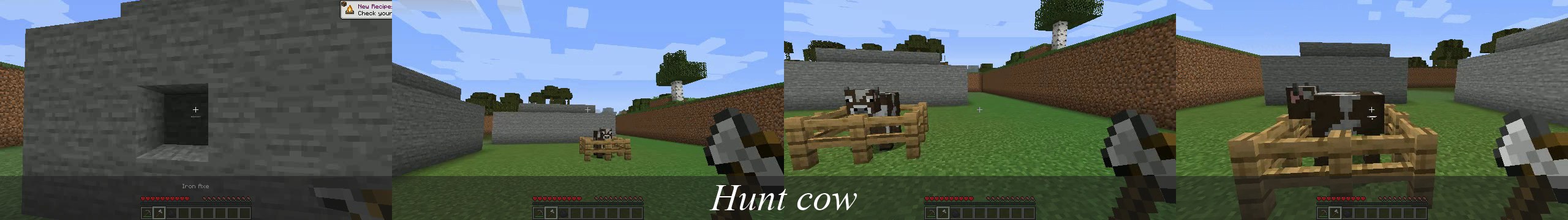}\\[2pt]
\includegraphics[width=\textwidth]{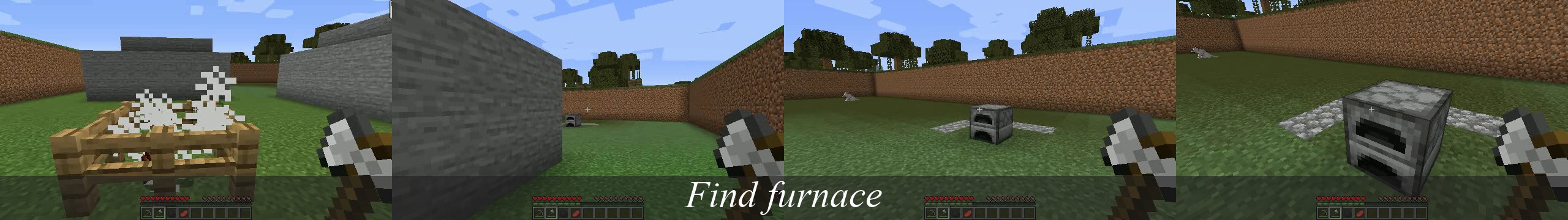}\\[2pt]
\includegraphics[width=\textwidth]{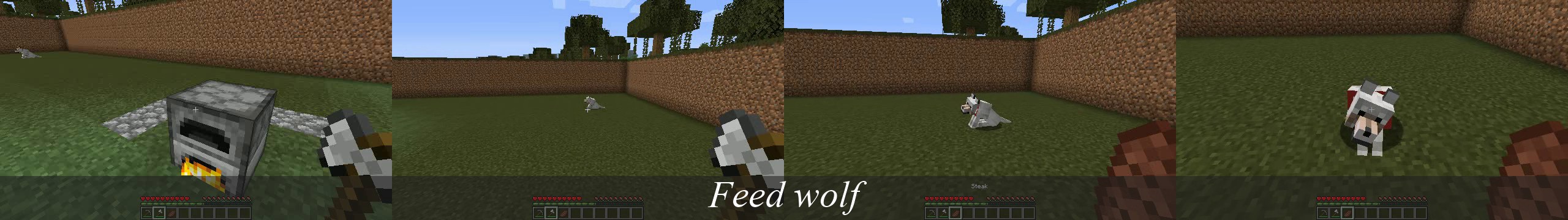}
\caption{Wolf Feeding chain (coal$\to$cow$\to$furnace$\to$wolf), shown in the same format as Figure~\ref{fig:chain-dpx}.}
\label{fig:chain-ccfw}
\end{figure}

\begin{figure}[H]
\centering
\includegraphics[width=\textwidth]{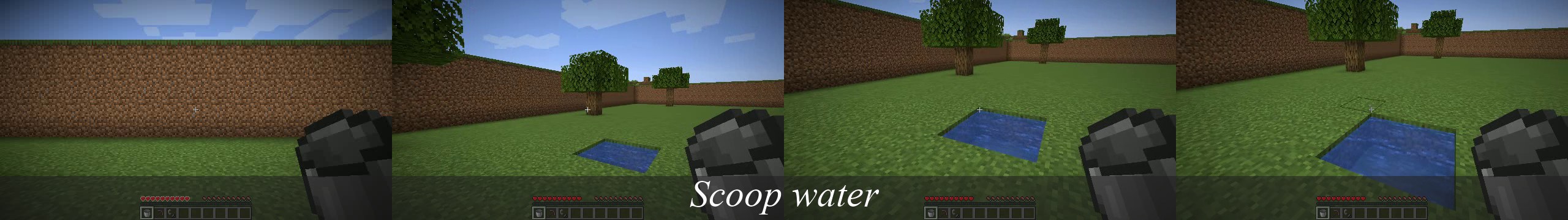}\\[2pt]
\includegraphics[width=\textwidth]{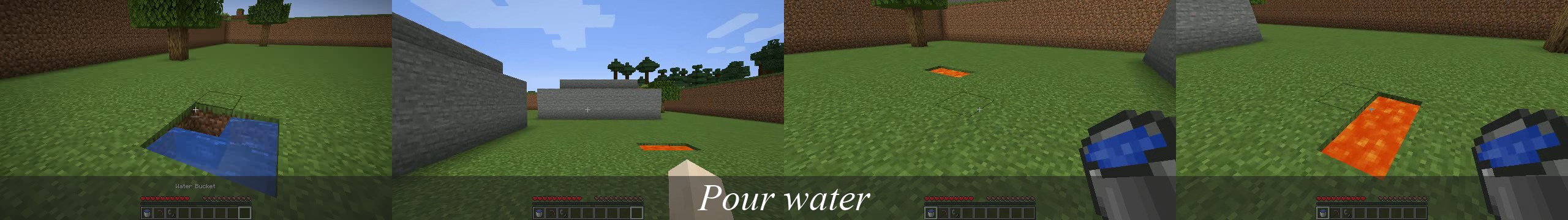}\\[2pt]
\includegraphics[width=\textwidth]{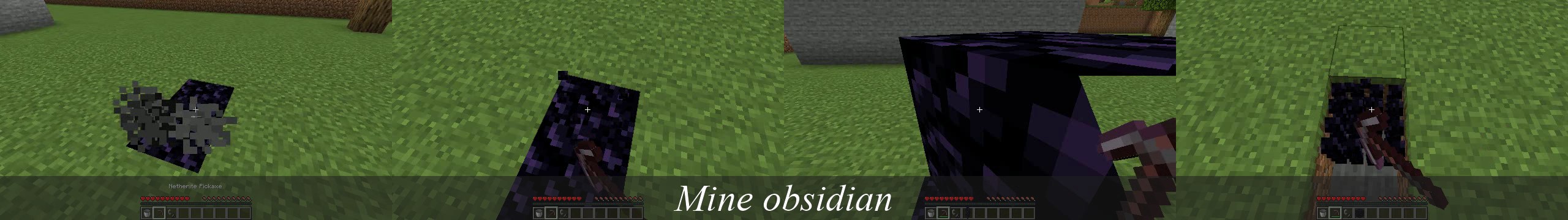}\\[2pt]
\includegraphics[width=\textwidth]{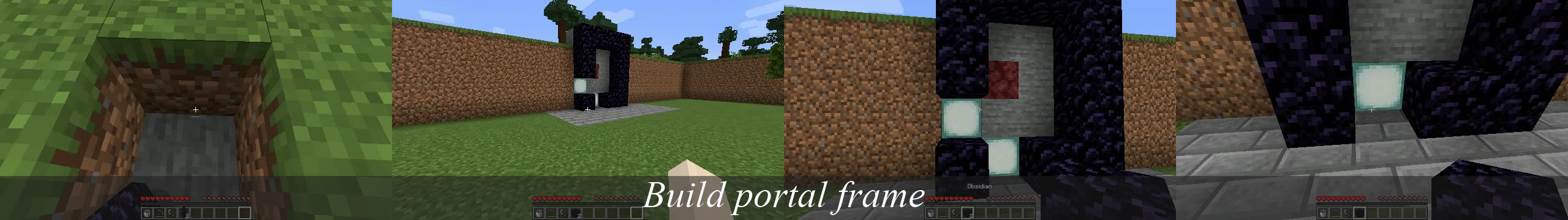}\\[2pt]
\includegraphics[width=\textwidth]{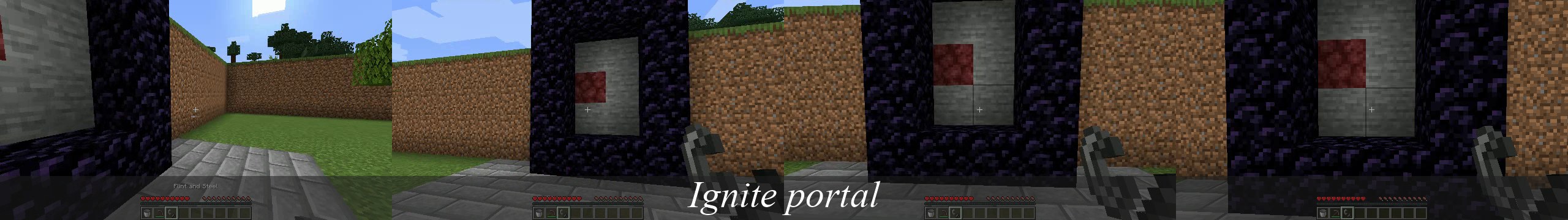}
\caption{Nether Portal chain (water$\to$lava$\to$obsidian$\to$portal$\to$ignition), shown in the same format as Figure~\ref{fig:chain-dpx}.}
\label{fig:chain-wlo}
\end{figure}

We evaluate three chains of increasing length, shown in Figures~\ref{fig:chain-dpx}, \ref{fig:chain-ccfw}, and~\ref{fig:chain-wlo}. Diamond Pickaxe requires mining an oak log, mining three diamond ores, and opening a placed crafting table. Wolf Feeding requires mining coal, hunting a cow, opening a furnace, and feeding cooked beef to a tamed wolf. Nether Portal requires scooping water, pouring it at a marked site, mining the resulting obsidian, and completing and igniting a portal frame. Each chain uses one fixed world and 50 paired policy seeds per method, and no method is trained on these chains.

\paragraph{Stage controller and helpers.}
A fixed stage controller switches the goal once the current stage is completed and attributed to the policy, and no language plan is generated. The controller also supplies stage quotas such as the three diamond ores, because the goal image does not encode counts. The world state, agent position, and camera orientation carry over without teleportation or forced camera alignment. GUI macros craft sticks after the oak stage, cook beef after the policy opens the furnace, and move the cooked beef to the hotbar. When the next stage needs a different item in hand, a helper equips it right after the previous stage is completed. Diamond Pickaxe scoring ends when the policy opens the existing table.

\section{Baseline Implementation}\label{app:baselines}
All baselines except the released ROCKET-2 checkpoint are trained on our demonstrations, starting from their released weights and using their native recipes and conditioning interfaces. STEVE-1~\papercite{NEURIPS2023_dd03f856} is evaluated with text goals and with video goals, GROOT~\papercite{ICLR2024_16986b69} with reference videos, and JARVIS-VLA~\papercite{li-etal-2025-jarvis} with text instructions. ROCKET-1~\papercite{Cai_2025_CVPR} and ROCKET-2~\papercite{cai2025rocket2} receive a current-view goal, built online by pointing at the target in the agent's observation with Molmo2-4B~\papercite{Clark_2026_CVPR} and segmenting it with SAM~2~\papercite{ICLR2025_45c1f6a8}. These models are part of the baseline pipelines and are not used by \attacca{}. ROCKET-2 is also evaluated with different-world goal images, both as the released checkpoint and after training on our dataset.

\providecommand{\GoalSpecDir}{figures/goal_spec}
\newlength{\GSw}\setlength{\GSw}{1.28in}          %
\newlength{\GSvw}\setlength{\GSvw}{0.4133in}      %
\newlength{\GSsw}\setlength{\GSsw}{0.48in}       %
\newcommand{\GSimg}[1]{\includegraphics[width=\GSw]{\GoalSpecDir/#1}}
\newcommand{\GSvid}[2]{%
  \ifx#2s\includegraphics[width=\GSvw]{\GoalSpecDir/video_#1_steve1_16_i000}\hspace{0.02in}%
    \includegraphics[width=\GSvw]{\GoalSpecDir/video_#1_steve1_16_i008}\hspace{0.02in}%
    \includegraphics[width=\GSvw]{\GoalSpecDir/video_#1_steve1_16_i015}%
  \else\includegraphics[width=\GSvw]{\GoalSpecDir/video_#1_groot_128_i000}\hspace{0.02in}%
    \includegraphics[width=\GSvw]{\GoalSpecDir/video_#1_groot_128_i064}\hspace{0.02in}%
    \includegraphics[width=\GSvw]{\GoalSpecDir/video_#1_groot_128_i127}\fi}
\newcommand{\GSobs}[2]{%
  \begin{tikzpicture}
    \node[inner sep=0pt] (i) {\GSimg{#1}};
    \node[anchor=north west, inner sep=1.2pt, fill=black, fill opacity=0.6, text opacity=1,
          text=white, font=\footnotesize] at (i.north west) {$o_{#2}$};
  \end{tikzpicture}}
\newcommand{\GStxt}[1]{\parbox[c]{\dimexpr\GSw-0.06in\relax}{\raggedright\itshape
  \ifdefined\microtypesetup\microtypesetup{protrusion=false}\fi ``#1''}}
\newcommand{\GSquery}[1]{\parbox[c]{\GSw}{\centering Molmo query\\[1pt]\itshape
  \ifdefined\microtypesetup\microtypesetup{protrusion=false}\fi ``#1''}}
\newcommand{\GSs}[2]{\parbox[t]{\GSsw}{\centering\includegraphics[width=\GSsw]{\GoalSpecDir/strip_#1_#2}\\[-0.5pt]%
  \fontsize{6}{6.6}\selectfont\GSname{#2}}}
\newcommand{\GSname}[1]{\csname GSn@#1\endcsname}
\newcommand{\GSdag}{\raisebox{0.5ex}{\fontsize{6}{6}\selectfont\textdagger}}
\newcommand{\GSrow}[1]{\raisebox{0.105in}{\textbf{#1}}}
\makeatletter
\@namedef{GSn@coal_ore}{coal ore}\@namedef{GSn@iron_ore}{iron ore}\@namedef{GSn@gold_ore}{gold ore}
\@namedef{GSn@lapis_ore}{lapis ore}\@namedef{GSn@diamond_ore}{diamond ore}\@namedef{GSn@emerald_ore}{emerald ore}
\@namedef{GSn@redstone_ore}{redstone ore}\@namedef{GSn@brain_coral_block}{brain coral\\block}
\@namedef{GSn@fire_coral_block}{fire coral\\block}\@namedef{GSn@horn_coral_block}{horn coral\\block}
\@namedef{GSn@cow}{cow}\@namedef{GSn@pig}{pig}\@namedef{GSn@chicken}{chicken}
\@namedef{GSn@white_sheep}{white sheep}\@namedef{GSn@gray_sheep}{gray sheep}
\@namedef{GSn@light_gray_sheep}{light gray\\sheep}\@namedef{GSn@brown_sheep}{brown sheep}
\@namedef{GSn@polar_bear}{polar bear}\@namedef{GSn@donkey}{donkey}\@namedef{GSn@panda}{panda}
\@namedef{GSn@oak_log}{oak log}\@namedef{GSn@birch_log}{birch log}\@namedef{GSn@bookshelf}{bookshelf}
\@namedef{GSn@hay_block}{hay block}\@namedef{GSn@pumpkin}{pumpkin}\@namedef{GSn@melon}{melon}
\@namedef{GSn@mossy_cobblestone}{mossy\\cobblestone}\@namedef{GSn@red_sandstone}{red\\sandstone}
\@namedef{GSn@dead_brain_coral_block}{dead brain\\coral block}\@namedef{GSn@green_terracotta}{green\\terracotta}
\makeatother

\expandafter\ifx\csname ifGoalSpecStrip\endcsname\relax
  \expandafter\newif\csname ifGoalSpecStrip\endcsname \GoalSpecStriptrue
\fi
\expandafter\ifx\csname ifGoalSpecPlaceObs\endcsname\relax
  \expandafter\newif\csname ifGoalSpecPlaceObs\endcsname \GoalSpecPlaceObstrue
\fi

\makeatletter
\@ifpackageloaded{float}{\def\GS@figplace{H}}{\def\GS@figplace{!ht}}
\edef\GoalSpecBeginFigure{\noexpand\begin{figure}[\GS@figplace]}
\makeatother

\clearpage
\subsection{Goal Specification per Method}\label{app:goal-spec}
All methods are evaluated on the same task instances, and the only difference in their inputs is how the target is specified (Figure~\ref{fig:goal-spec}).
Text-goal policies receive an instruction naming the target class, and STEVE-1 receives a comma-separated list of equivalent commands, which its text prior maps to a visual goal embedding.
Video-goal policies receive a clip of a human completing the same task on the same class (16 frames for STEVE-1 and 128 for GROOT), taken from demonstrations not used for training.
\mbox{ROCKET-1} and \mbox{ROCKET-2} with Molmo receive no separate goal. Every 30 steps, Molmo2-4B is asked to ``Point to $\langle$class$\rangle$'' in the agent's current view, SAM~2 segments the object at the returned point, and the view with this mask becomes the goal, which \mbox{ROCKET-2} holds until the next query and \mbox{ROCKET-1} tracks at every step.
ROCKET-2$^{\dagger}$, ROCKET-2, and \attacca{} receive the same goal image, taken in a different world, with its target mask\ifGoalSpecStrip{} (Figure~\ref{fig:goal-images})\fi.

\GoalSpecBeginFigure
\centering
\ifdefined\microtypesetup\microtypesetup{protrusion=false}\fi %
\scriptsize
\setlength{\tabcolsep}{0pt}
\renewcommand{\arraystretch}{1.0}
\begin{tabular}{@{}>{\raggedright\arraybackslash}m{0.78in}@{\hspace{0.04in}}>{\raggedright\arraybackslash}m{0.64in}@{\hspace{0.05in}}m{\GSw}@{\hspace{0.05in}}m{\GSw}@{\hspace{0.05in}}m{\GSw}@{}}
\toprule
\textbf{Goal modality} & \textbf{Policy} & \textbf{Mine}: coal ore & \textbf{Hunt}: cow & \textbf{Place}: on oak log \\
\midrule
Text & STEVE-1 & \GStxt{mine coal ore, dig coal ore, collect coal ore, get coal ore} & \GStxt{kill the cow, hunt the cow, attack the cow} & \GStxt{place a block on the oak log, put a block on the oak log} \\
\addlinespace[3pt]
 & JARVIS-VLA & \GStxt{mine coal ore} & \GStxt{hunt cow} & \GStxt{place a block on the oak log} \\
\cmidrule{1-5}
Video & STEVE-1 & \GSvid{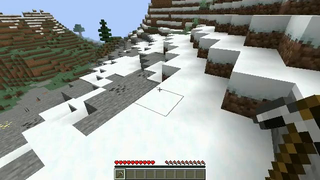}{s} & \GSvid{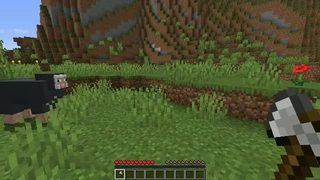}{s} & \GSvid{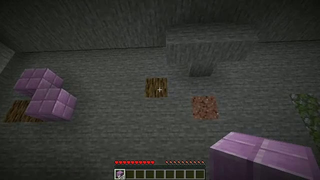}{s} \\
\addlinespace[2pt]
 & GROOT & \GSvid{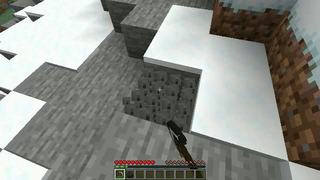}{g} & \GSvid{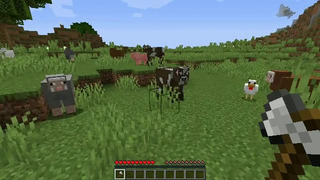}{g} & \GSvid{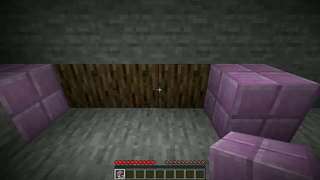}{g} \\
\cmidrule{1-5}
Current view {\footnotesize$o_t$}\newline $+$ Molmo $+$ SAM & ROCKET-1\newline ROCKET-2 & \GSobs{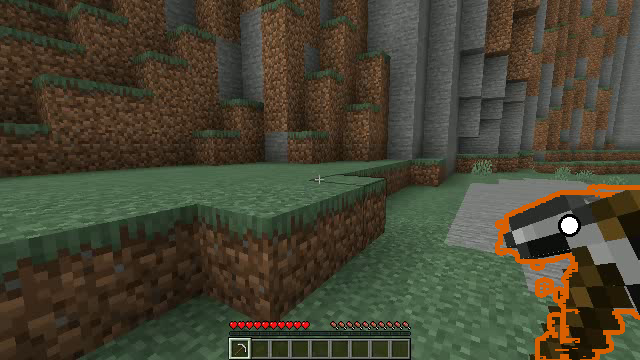}{0} & \GSobs{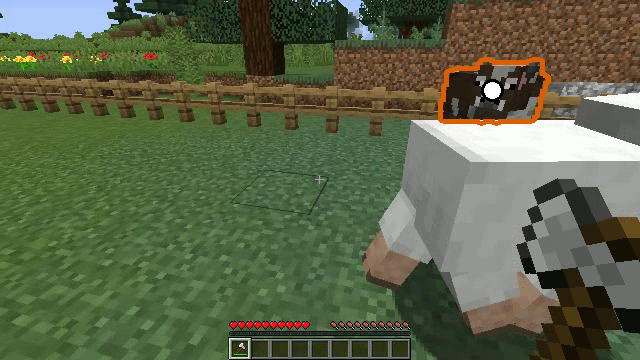}{150} & \ifGoalSpecPlaceObs\GSobs{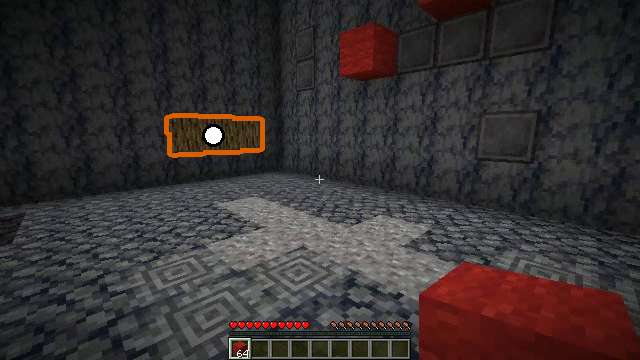}{90}\else\GSquery{Point to oak log}\fi \\
\cmidrule{1-5}
Different world\newline image $+$ mask & ROCKET-2\GSdag\newline ROCKET-2\newline \textbf{\attacca{} (Ours)} & \GSimg{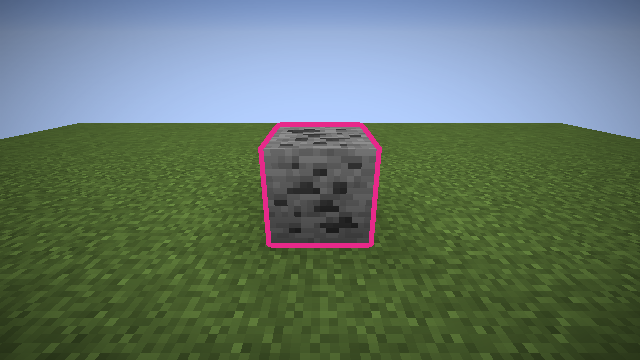} & \GSimg{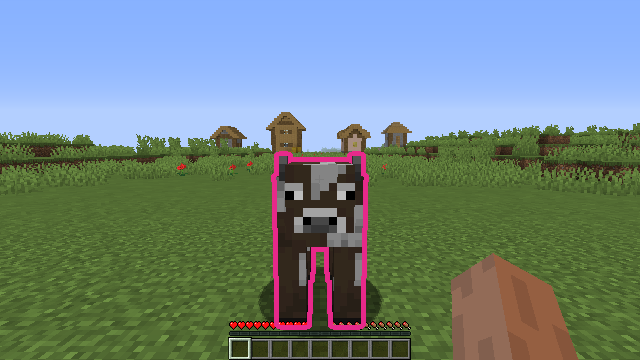} & \GSimg{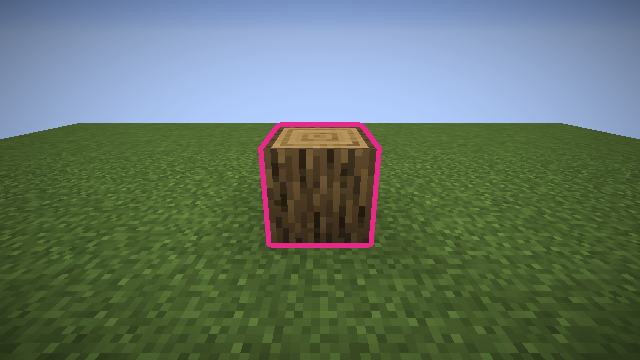} \\
\bottomrule
\end{tabular}
\caption{Goal given to each method for one target class per task. Video goals are shown by the first, middle, and last frame of the clip. Current-view goals show the agent's observation $o_t$ at a Molmo query step, with the Molmo point (white dot) and the SAM~2 mask (orange outline) of the ROCKET-2 pipeline. At $t{=}0$ in Mine the target is not yet in view, and the point falls on the held pickaxe\ifGoalSpecPlaceObs\else. For Place, only the Molmo query is shown\fi. Different-world goals show the goal image with its target mask (magenta outline).}
\label{fig:goal-spec}
\ifGoalSpecStrip
\vspace{4pt}
\begin{tabular}{@{}>{\raggedright\arraybackslash}p{0.33in}*{6}{c@{\hspace{0.025in}}}c@{\hspace{0.085in}}c@{\hspace{0.025in}}c@{\hspace{0.025in}}c@{}}
 & \multicolumn{7}{c}{ID (seen in training)} & \multicolumn{3}{c}{OOD (held out)} \\
\cmidrule(r{0.04in}){2-8}\cmidrule(l{0.02in}){9-11}
\GSrow{Mine} & \GSs{mine}{coal_ore} & \GSs{mine}{iron_ore} & \GSs{mine}{gold_ore} & \GSs{mine}{lapis_ore} & \GSs{mine}{diamond_ore} & \GSs{mine}{emerald_ore} & \GSs{mine}{redstone_ore} & \GSs{mine}{brain_coral_block} & \GSs{mine}{fire_coral_block} & \GSs{mine}{horn_coral_block} \\
\addlinespace[3pt]
\GSrow{Hunt} & \GSs{hunt}{cow} & \GSs{hunt}{pig} & \GSs{hunt}{chicken} & \GSs{hunt}{white_sheep} & \GSs{hunt}{gray_sheep} & \GSs{hunt}{light_gray_sheep} & \GSs{hunt}{brown_sheep} & \GSs{hunt}{polar_bear} & \GSs{hunt}{donkey} & \GSs{hunt}{panda} \\
\addlinespace[3pt]
\GSrow{Place} & \GSs{place}{oak_log} & \GSs{place}{birch_log} & \GSs{place}{bookshelf} & \GSs{place}{hay_block} & \GSs{place}{pumpkin} & \GSs{place}{melon} & \GSs{place}{mossy_cobblestone} & \GSs{place}{red_sandstone} & \GSs{place}{dead_brain_coral_block} & \GSs{place}{green_terracotta} \\
\end{tabular}
\caption{Goal images of all short-horizon target classes, cropped around the target, with the target mask outlined in magenta. Mine and Place goals share one camera pose and target position.}
\label{fig:goal-images}
\fi
\end{figure}

\clearpage
\section{Additional Failure Analysis}\label{app:failure-analysis}
\cref{fig:failure-diamond} shows the diamond scenario of \cref{sec:exemplar-execution} for three goal specifications, and \cref{fig:failure-oak} shows the oak scenario in the same format. Row (a) uses the current-view goal, which Molmo and SAM~2 construct from the agent's observation every 30 steps. At $o_{1}$ the diamond is out of view, so Molmo finds no target and no goal is built. The diamond enters the view at $o_{121}$ and the policy approaches it, but at $o_{151}$ the mask covers the whole stone structure instead of the ore, and later points land on UI elements. Online goal construction therefore cannot solve search when the target starts out of view.

Row (b) repeats the fine-tuned ROCKET-2 rollouts of \cref{fig:failure-rocket2}. Row (c) shows \attacca{} with the bottom goal image of (b) and no external grounding module. It switches from Search to Approach at $o_{87}$, when the predicted mask first activates, begins interacting at $o_{182}$, and mines the third diamond ore by $o_{264}$. The oak scenario shows the same failure modes. The online pipeline cannot build a goal while the log is out of view, ROCKET-2 aligns its view with a same-world goal instead of interacting and loses the target under a different-world goal, and \attacca{} completes the task.
\begin{figure}[H]
\centering
\includegraphics[width=\textwidth]{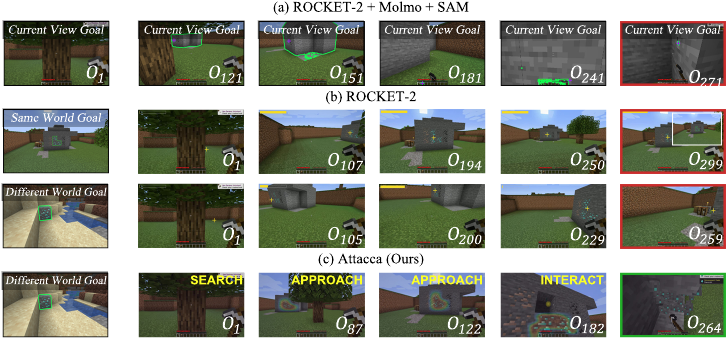}
\caption{Diamond scenario. (a) Current-view goal with Molmo points (purple) and SAM~2 masks (green). (b) ROCKET-2 with same-world (top) and different-world (bottom) goals, with its predicted point and visibility. (c) \attacca{} with the bottom goal image of (b) and its predicted phase (yellow).}
\label{fig:failure-diamond}
\end{figure}

\begin{figure}[H]
\centering
\includegraphics[width=\textwidth]{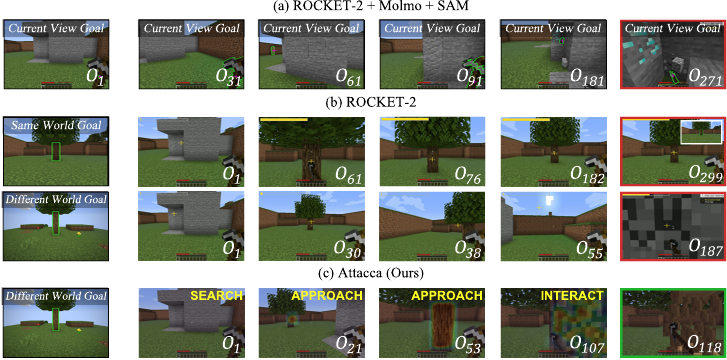}
\caption{Failure analysis on the mine-oak-log task, following the format of Figure~\ref{fig:failure-diamond}.}
\vspace{-0.1in}
\label{fig:failure-oak}
\end{figure}

\section{Limitations}\label{app:limitations}
This work focuses on the low-level policy of a hierarchical embodied agent and improves its ability to carry execution from one task to the next under state continuity. Our evaluation therefore isolates goal-directed embodied control in Minecraft and leaves the other parts of a long-horizon task to the evaluation system. A scripted stage controller issues the goal image of each task and detects its completion, and helper functions handle items at task transitions. For example, when the agent finishes cooking beef at the furnace, a macro moves the cooked beef to the hotbar and a helper equips it once the beef is cooked, so the agent only needs to find the wolf and feed it.

These parts correspond to the role of the high-level planner, and connecting \attacca{} to a planner would take them over. As in existing hierarchical agents, the planner would decide when each subgoal is achieved and dispatch the next one. The remaining interface is turning a language subgoal into a masked goal image. Since \attacca{} is trained with goal images from other worlds, this image does not need to come from the execution world and can be retrieved from a fixed library of goal images or built by segmenting the target in any image that contains it. We have not yet evaluated such an integrated system and leave it to future work. Finally, all of our experiments are conducted in Minecraft. Extending \attacca{} to other embodied domains, such as robotic navigation and manipulation, is another direction for future work.

\end{document}